\documentclass[twoside,12pt]{article}

\usepackage{jmlr2e}
\usepackage{animate}
\usepackage{longtable}
\usepackage{enumitem}
\usepackage{subcaption}
\usepackage{amsmath}
\usepackage{physics}
\usepackage{bm}
\usepackage{cleveref}
\usepackage{mathrsfs}
\usepackage{multirow}
\usepackage{float}
\usepackage{mathtools,amssymb,amsmath}
\usepackage{amsfonts}
\RequirePackage{bm}
\usepackage[linesnumbered,ruled]{algorithm2e}
\usepackage{algpseudocode}
\usepackage{multirow, makecell}
\usepackage{enumitem}
\usepackage{lipsum}
\usepackage[table]{xcolor}

\definecolor{orangec}{RGB}{255,158,62}

\newcommand{\mb}[1]{\boldsymbol{\mathbf{#1}}}

\newcommand{\wh}[1]{\widehat{#1}}

\ShortHeadings{Unified Transfer Learning Models}{Unified Transfer Learning Models}
\firstpageno{1}

\begin{document}
	
	\title{Unified Transfer Learning Models in High-Dimensional Linear Regression}
	
	\author{\name Shuo Shuo Liu \email shuoliu.academic@gmail.com  
	\\
	\addr 
	Columbia University
\\}
	
	
	\maketitle
	
	\begin{abstract}
Transfer learning plays a key role in modern data analysis when: (1) the target data are scarce but the source data are sufficient; (2) the distributions of the source and target data are heterogeneous.
This paper develops an interpretable unified transfer learning model, termed as UTrans, which can detect both transferable variables and source data.
More specifically, we establish the estimation error bounds and prove that our bounds are lower than those with target data only.
Besides, we propose a source detection algorithm based on hypothesis testing to exclude the nontransferable data.
We evaluate and compare UTrans to the existing algorithms in multiple experiments.
It is shown that UTrans attains much lower estimation and prediction errors than the existing methods, while preserving interpretability.
We finally apply it to the US intergenerational mobility data and compare our proposed algorithms to the classical machine learning algorithms.
	\end{abstract}
	
	
	\section{Introduction}
	\label{sec: intro}
	Predictive models, which employ the training data to make predictions, have been effectively used to guide decision making in various applications.
Modern data extraction techniques further improve model performance and statistical inference by utilizing a collection of massive and diverse data \citep{zhuang2020comprehensive, tripuraneni2020theory, liu2023adaptive}.
With data collected from multiple sources, the superior predictive ability of these models relies on the hypothesis that these multi-source data share a homogeneous or similar  distribution.
When such hypothesis fails, most predictive models using the training data lose the prediction power and require reconstruction by gathering new data from the same distribution.
However, the cost of collecting new data or the privacy limit of integrating multiple data may hinder the reconstruction.
To improve the predictive performance, one of the possible solutions is to transfer and integrate the useful source data. 
In this scenario, transferring data knowledge from one source (namely, \textit{source data}) to another (namely, \textit{target data}) would be required, of which the learning process is called \textit{transfer learning} in the literature \citep{olivas2009handbook}.
The three main themes for researchers in transfer learning are: what to transfer, when to transfer, and how to transfer? 

Transfer learning has drawn extensive attention for decades and been applied in many fields including Web-document classification, Wifi data calibration, medical diagnosis, and so on.
See more examples in the recent survey paper \citet{zhuang2020comprehensive}.
Beyond these applications of transfer learning in the machine learning community, some methodological and theoretical works are also developed.
\citet{yogatama2014efficient} proposes a fast and effective algorithm for automatic hyperparameter tuning that utilizes sequential model-based optimization (SMBO) to construct a common response surface across datasets, enabling generalization.
\citet{wei2018transfer} studies how to automatically determine what and how to transfer by leveraging previous transfer learning experiences.
\citet{bellot2019boosting} introduces a survival prediction model that enhances predictions in a small data domain, like a local hospital, by leveraging related data from other domains, constructing an ensemble of weak survival predictors that iteratively adapts marginal distributions to improve predictions for target patients of interest.
\citet{tripuraneni2020theory} studies a two-stage empirical risk minimization procedure to transfer learning and provides generalization bounds with general losses, tasks, and features.
However, little attention has been paid to interpretable transfer learning in the statistical framework, which can generate interpretable results and study the corresponding theoretical properties.
In this paper, we aim to fill this gap, develop new statistical transfer learning models in the context of high-dimensional data, and improve the predictive performances of the existing transfer learning models.

\subsection{High-dimensional transfer learning models}

High-dimensional linear models based on one source data with suitable regularizations have been developed extensively over the past decade \citep{tibshirani1996regression, FanLi} due to the high-dimensional nature of real-world data.
For example, in gene expression data, it is common to encounter a few observations but hundreds of thousands of genes.
In financial data, it is widely seen that the number of features is much larger than the number of individual stocks. 
The high-dimensional linear regression model, with single-source data, takes the form
$\mb y_1=\mb X_1\mb\beta_1+\mb\epsilon_1,$
where $\mb y_1\in\mathbb{R}^{n_1}$, $\mb X_1\in\mathbb{R}^{n_1\times p}$, $\mb\beta_1\in\mathbb{R}^{p}$, and $\mb\epsilon_1\in\mathbb{R}^{n_1}$.
With the high-dimensional data, we allow the dimension $p\gg n_1$ for the unknown coefficient vector $\mb\beta_1$.

Transfer learning has been studied recently in statistical models \citep{Li2022, tian2022transfer, lin2022transfer}.
For example, in the high-dimensional linear regression model \citep{Li2022}, the target model is
$y_{0i}=\mb x_{0i}^\top \mb\beta_0+\epsilon_{0i}, i=1,\cdots, n_0$
and the source model from the $k$-th source data, $k=1,\cdots,K'$, is 
$y_{ki}=\mb x_{ki}^\top \mb\beta_k+\epsilon_{ki}, i=1,\cdots, n_k,$
where $\mb x_{ki}\in\mathbb{R}^{p}$ and $\mb\beta_k\in\mathbb{R}^{p}, k=0,1,\cdots,K'$.
Useful source data are transferred to the target data only if the transferring set $\mathcal{A}_h$ satisfies
$\mathcal{A}_h=\{1\leq k\leq K': \|\mb\beta_0-\mb\beta_k\|_q\leq h\}$
for a relatively small \textit{transferring level} $h$.
This model, named Trans-Lasso, leverages the linear regression model to bridge the source and target data and transfers source data to the target data when $k\in\mathcal{A}_h$.
Trans-Lasso solves $\mb w$ from the source data in the first step and then debiases the estimation from the target data in the second step.
Let $n_{\mathcal{A}_h}$ denote the sample size of the source data in $\mathcal{A}_h$.
More specifically, the first step solves 
$$\hat{\mb w}=\underset{\mb w \in \mathbb{R}^{p}}{\arg \min }\left\{\frac{1}{2n_{\mathcal{A}_h}} \sum_{k \in \mathcal{A}_h}\left\|\mb y^{(k)}-\mb X^{(k)} \mb w\right\|_{2}^{2}+\lambda\|\mb w\|_{1}\right\}$$
via integrating the diverse information from multiple sources.
\citet{tian2022transfer} and \citet{lin2022transfer} extend the results of \citet{Li2022} to high-dimensional generalized linear models (GLMs) and functional linear models, respectively. 
Some consistent estimators of $\mathcal{A}_h$ are required, such as the Q-aggregation \citep{Li2022} and data-splitting estimator under some conditions \citep{tian2022transfer}.
Other nonparametric predictive models also exist in the literature, such as the adaptive transfer learning with minimax optimal rates of convergence based on $k$-nearest neighbour \citep{cai2021transfer, reeve2021adaptive}.
Noteworthy, multi-task learning is a closely related topic to transfer learning, but with different goals and interests.
Multi-task learning method integrates multiple learning tasks simultaneously, while exploiting a shared structure across all tasks.
For example, see the structure of Data Shared Lasso \citep{gross2016data, ollier2017regression} for high-dimensional multi-task learning.
In contrast, the interest of transfer learning is to learn the target data only by transferring some shared knowledge from the source data.
Therefore, learning the source data is not the focus of transfer learning.

In this paper, our contributions include
\begin{enumerate}
	\item We propose a novel unified transfer learning model by redefining the design matrix and the response vector in the context of the high-dimensional linear regression with a flexible penalty function.
	When the transferring set is known, the theoretical results show that it attains tighter upper bounds of the $\ell_1/\ell_2$ estimation errors than Lasso using the target data only.
    We also compare our theoretical results to the existing methods.
	\item Detecting the transferable data, including transferable source data and transferable variables, is a major task in transfer learning.
	Our unified model is able to automatically identify the transferable variables after model estimation.
	To the best of our knowledge, this is the first work for identifying the transferable variables by the model's nature and the first work for detecting transferable source data by hypothesis testing.
\end{enumerate}
	
	\section{Unified Transfer Learning Models}
	\label{sec: unified}
	\textbf{Notations: }We denote scalars with unbolded letters (e.g.,  sample size $n$ and dimensionality $p$), (random) vectors with boldface lowercase letters (e.g.,  $\mb y$ and $\mb\beta$),  and matrices with boldface capital letters (e.g., $\mb X$). 
Let $\{(\mb X_k, \mb y_k): \mb X_k\in\mathbb{R}^{n_k\times p},  \mb y_k\in\mathbb{R}^{n_k}\}_{k=1}^{K'}$ denote the multiple source data and let $(\mb X_0, \mb y_0)$ be the target data. 
We use $\top$ to represent the transpose of vectors or matrices, such as $\mb x^\top$ and $\mb X^\top$.
For a $p$-dimensional vector $\mb x=(x_1, \cdots, x_p)$, the $\ell_0$ norm is the number of non-zero elements.
$\|\mb x\|_q$ and $\|\mb x\|_{\infty}$ are the $\ell_q$ norm and maximum norm, respectively.
$|\mathcal{M}|$ denotes the cardinality of the set $\mathcal{M}$.
A set with superscript $c$ denotes its complement.
We use letters $C$ and $c$ with different subscriptions to denote the positive and absolute constants.
Let $a_n=O(b_n)$ denote $|a_n/b_n|\leq c$ for some constant $c$ when $n$ is large enough.
Let $a_n=O_P(b_n)$ and $a_n\lesssim b_n$ denote $P(|a_n/b_n|\leq c)\rightarrow 1$ for $c<\infty$.
Let $a_n=o_P(b_n)$ denote $P(|a_n/b_n|> c)\rightarrow 0$ for $c>0$.
Finally, $a_n\asymp b_n$ means that $a_n/b_n$ converges to some positive constant.

Throughout the following sections,
we abbreviate $\mathcal A_h$ by $\mathcal A$ for simplicity and use $K$ to denote the number of transferable source data.
The first step (namely, transferring step) of the transfer learning models for high-dimensional linear regression in \citet{Li2022} 
is essentially equivalent to stacking all source data, assuming $\mathcal{A}$ is known:
\begin{equation}
	\label{eq:stacking}
	\hat{\mb w}=\underset{\mb w \in \mathbb{R}^{p}}{\arg \min }\left\{\frac{1}{2n_{\mathcal{A}}}\left\|\mb y'-\mb X' \mb w\right\|_{2}^{2}+\lambda\|\mb w\|_{1}\right\},
\end{equation}
where $\mb y'=[\mb y_1^\top,\cdots, \mb y_K^\top]^\top$, $\mb X'=[\mb X_1^\top,\cdots, \mb X_K^\top]^\top$, and $n_{\mathcal{A}}$ is the total sample size of the source data.
\citet{tian2022transfer} proposes to stack the source data and the target data in the GLMs in the transferring step.
We call these methods as vertical stacking methods.
The assumption behind these methods is that the data (the source data in $\mathcal{A}$ or the target and the source data in $\mathcal{A}$) share a similar coefficient $\mb w$.
Stacking the data in the way of Eq. (\ref{eq:stacking}) may produce a better estimation when different data are close, but might be insufficient to identify the transferable variables in the source data.
For example, we are unable to identify the transferable variables to the target data for the $k$-th source data.
Therefore, we consider a new approach, unified transfer learning models, for transfer learning in the high-dimensional linear regression in this section.

\subsection{$\mathcal A$-UTrans: transfer learning with known $\mathcal A$}
Instead of stacking the target data and the source data in $\mathcal A$ vertically, we propose to stack them both vertically and horizontally by 
\begin{equation*}
	\left[\begin{array}{c}
		\mb y_{1} \\
		\vdots\\
		\mb y_K\\
		\mb y_{0}
	\end{array}\right]=\left[\begin{array}{c}
		\mb X_{1} \\
		\mb 0\\
		\vdots\\
		\mb 0
	\end{array}\right]\left(\mb\beta_{1}-\mb\beta_{0}\right)+\cdots+\left[\begin{array}{c}
		\mb X_{1} \\
		\vdots\\
		\mb X_K\\
		\mb X_{0}
	\end{array}\right] \mb\beta_{0}+\mb\epsilon
\end{equation*}
where $\mb\epsilon=[\mb\epsilon_{1}^\top, \cdots, \mb\epsilon_{K}^\top, \mb\epsilon_{0}^\top]^\top$.
The aforementioned model can be written as $\mb y=\mb X\mb\beta+\mb\epsilon$, where 
$\mb y=[\mb y_1^\top,\cdots, \mb y_K^\top, \mb y_0^\top]^\top$, $\mb\beta=[(\mb\beta_1-\mb\beta_0)^\top, (\mb\beta_2-\mb\beta_0)^\top, \cdots, \mb\beta_0^\top]^\top\in\mathbb{R}^{p^*}$, and
\begin{equation}
	\label{eq: X}
	\mb X=\begin{bmatrix}
		\mb X_1 & \mb 0 & \cdots&\cdots&\mb 0&\mb X_1\\
		\mb 0 & \mb X_2 & \mb 0&\cdots&\mb 0&\mb X_2\\
		\vdots&\vdots&\vdots&\vdots&\vdots&\vdots\\
		\mb 0& \cdots&\cdots&\cdots&\mb X_K&\mb X_K\\
		\mb 0& \cdots&\cdots&\cdots&\mb 0&\mb X_0\\
	\end{bmatrix}\in\mathbb{R}^{(n_{\mathcal A}+n_0)\times p^*}
\end{equation}
where $p^*=Kp+p$ and $\mathcal{A}=\{k:1\leq k\leq K\}$.
In this paper, we assume that $K$ is fixed.

We consider a more general penalty function that includes the Lasso used in the current literature and other nonconvex regularizers to deal with the high-dimensional data.
The loss function of the penalized least square is
\begin{equation}
	\label{eq: loss}
	\mathcal{L}_n(\mb\beta)=\frac{1}{2\left(n_{\mathcal{A}}+n_{0}\right)}\|\mathbf{y}-\mathbf{X} \boldsymbol{\beta}\|_2^{2}+P_{\lambda}(\mb\beta)
\end{equation}
and we denote this unified transfer learning model as \textbf{$\mathcal A$-UTrans}.
We solve $\wh{\mb\beta}_0$ by the coordinate descent algorithm for nonconvex penalized regression \citep{breheny2011coordinate}.
Note that $\wh{\mb\beta}_0$ equals the last $p$ elements of $\wh{\mb\beta}$.

The penalty function $P_{\lambda}(\mb\beta)=\sum_{j=1}^{p^*} p_{\lambda}\left(\left|\beta_{j}\right|\right)$ satisfies the following conditions
\begin{itemize}
	\item (i) $P_{\lambda}(0)=0$ and $P_{\lambda}(t)$ is symmetric around 0.
	\item (ii) $P_{\lambda}(t)$ is differentiable for $t \neq 0$ and $\lim _{t \rightarrow 0^{+}} P_{\lambda}^{\prime}(t)=\lambda L$.
	\item (iii) $P_{\lambda}(t)$ is a non-decreasing function for $t \geq0$.
	\item (iv) $P_{\lambda}(t) / t$ is a non-increasing function for $t >0$.
	\item (v) There exists $\tau>0$ such that $P_{\lambda}(t)+\frac{\tau}{2} t^{2}$ is convex.
\end{itemize}

Conditions (i)--(iii) are relatively mild and used in \citet{zhang2012general}.
Condition (iv) makes sure that the bound of error $\|\hat{\mb\beta}-\mb\beta\|_2$ is vanishingly small.
These mild conditions on $P_{\lambda}(\mb\beta)$ are commonly satisfied by many regularizers including Lasso \citep{tibshirani1996regression}, SCAD \citep{FanLi}, and MCP \citep{zhang2010nearly}.
For more details, refer to \citet{Loh2015}.

We argue two benefits of our $\mathcal A$-UTrans models.
First, the unified transfer learning model explicitly writes the contrasts $\mb\beta_k-\mb\beta_0$.
The $k$-th source data are transferable if $\mb\beta_k-\mb\beta_0=\mb 0$.
This method, therefore, provides an opportunity to detect transferable source by testing $\mb\beta_k-\mb\beta_0=\mb 0$.
In Section \ref{sec: detect}, we propose to use hypothesis testing to detect transferable source data.
Second, other than detecting the transferable source data, our method can also detect the transferable variables in each source data.
For example, we obtain the set containing the transferable variables in the $k$-th source data by $\mathcal T_k=\{j: (\wh{\mb\beta_k-\mb\beta_0})_{j}=0, 1\leq j\leq p\}$.

\subsection{Theoretical properties of $\mathcal A$-UTrans}
We define the parameter space of $\mathcal A$-UTrans by $\Theta(s,h)$, which is
\begin{equation*}
	\left\{\mb\beta:  \max_{k\in\mathcal A} \|\mb\beta_k-\mb\beta_0\|_1\leq h, \|\mb\beta_0\|_0\leq s, \|\mb\beta_k-\mb\beta_0\|_0\leq Cs\right\}
\end{equation*}
for some constant $C$ and $s=\|\mb\beta_0\|_0$ is the sparsity level.
Note that this parameter space specifies the sparsity of $\mb\beta_0$ and constraints the maximum $\ell_1$ distance between $\mb\beta_k$ and $\mb\beta_0$ to $h$.
We further impose the following conditions to study the theories of $\mathcal A$-UTrans:

\begin{itemize}
	\item C1. Each row of $\mb X_k, k\in\mathcal{A}\cup \{0\},$ is independent and identically distributed (i.i.d) normal random vector with mean zero and covariance matrix $\mb\Sigma_k$. 
	\item C2. The random noises $\epsilon_{ki}$ in the $k$-th source data, $i=1, \cdots, n_k$ and $k\in\mathcal{A}\cup \{0\},$ are i.i.d sub-Gaussian random variable with mean zero and parameter $\sigma_k^2$.
	\item C3. The sample covariance matrix 
$\widehat{\mb\Sigma}_k=\frac{1}{n_k}\mb X_k^\top\mb X_k$, $k\in\mathcal{A}\cup \{0\}$, satisfies the restricted strong convexity (RSC) condition 
	$$
	\mb{\Delta}_k^{\top} \widehat{\boldsymbol{\Sigma}}_k \boldsymbol{\Delta}_k \geq v_k\|\boldsymbol{\Delta}_k\|_{2}^{2}-\tau_k \sqrt{\frac{\log p}{n_k}}\|\boldsymbol{\Delta}_k\|_{1}
	$$
	for any $\boldsymbol{\Delta}_k \in \mathbb{R}^{p} \text { and }\|\boldsymbol{\Delta}_k\|_{1} \geq 1$,
	where $v_k>0$ and $\tau_k\geq 0$. 
\end{itemize}

In C1, the source data and the target data are assumed to have Gaussian designs.
The covariance matrix $\mb\Sigma_k$ can be homogeneous or heterogeneous among the source and the target data.
Different from \citet{Li2022} whose theories are established separately with the homogeneous and heterogeneous covariance matrices, our theories can incorporate both.
This condition is for theoretical convenience and can be relaxed to sub-Gaussian random variable.
Condition C2 assumes the sub-Gaussian random noises for the source and the target data, which are used for the convergence rate analysis.
Condition C3 assumes the RSC condition for each sample covariance matrix. 
This condition is widely used to study the non-asymptotic error bounds in high-dimensional statistics.
It is shown that the RSC condition is met with high probability under sub-Gaussian assumption \citep{Alekh2012, Loh2015, liu2022multiple}.
We mention that the RSC condition can be replaced by the restricted eigenvalue (RE) condition \citep{bickel2009simultaneous, van2009conditions}.
For simplicity, denote $n=n_{\mathcal A}+n_0$.
We have the following RSC condition on the sample covariance matrix of $\mb X$.

\begin{theorem}
	\label{th: rsc}
	Let $\wh{\mb\Sigma}=\mb X^\top\mb X/n$ be the sample covariance matrix of $\mb X$.
    	With the RSC conditions on each $\wh{\mb\Sigma}_k$, we have	
    	\begin{equation*}
    		\label{eq: rsc2}
    	\wh{\mb\Delta}^\top\wh{\mb\Sigma}\wh{\mb\Delta}\geq v'\|\wh{\boldsymbol{\Delta}}\|_{2}^{2}-\tau_0 \left(\sqrt{\frac{n_m\log p}{n^2}}+\sqrt{\frac{n_0\log p}{n^2}}\right)\|\wh{\boldsymbol{\Delta}}\|_{1} 
    	\end{equation*}
    for $\wh{\mb\Delta}=\wh{\mb\beta}-\mb\beta \in \mathbb{R}^{p^*}$, where $v'=\min_k v_k n_k/n>0$, $\tau_0=\max_k \tau_k (K+1)\geq 0$, and $n_m=\max_{k\in\mathcal A}n_k$, $k\in\mathcal A$.
\end{theorem}

Theorem \ref{th: rsc} implies that the sample covariance matrix $\wh{\mb\Sigma}$ in the unified model admits a similar RSC condition as that from a single source data.
The term $\sqrt{n_m\log p/n^2}+\sqrt{n_0\log p/n^2}\lesssim \sqrt{\log p/n}$ in the lower bound is essential for establishing the estimation error bound.
Note that this term is upper bounded by $\sqrt{\log p/n_0}$.
Thus, a tighter error bound than the model using target data only can be established.
From this theorem, we observe
$$\wh{\mb\Delta}^\top\wh{\mb\Sigma}\wh{\mb\Delta}\geq v'\|\wh{\boldsymbol{\Delta}}\|_{2}^{2}-2\tau_0 \sqrt{\frac{\log p}{n}}\|\wh{\boldsymbol{\Delta}}\|_{1},$$
which trivially holds for $\frac{\|\wh{\boldsymbol{\Delta}}\|_{1}}{\|\wh{\boldsymbol{\Delta}}\|_2^2}\geq \frac{v'}{2\tau_0}\sqrt{\frac n{\log p}}$ since the left-hand side is nonnegative.
Thus, we only enforce a type of strong convexity condition over a cone of the form 
$$\left\{\frac{\|\wh{\boldsymbol{\Delta}}\|_{1}}{\|\wh{\boldsymbol{\Delta}}\|_2^2}\leq \frac{v'}{2\tau_0}\sqrt{\frac n{\log p}}\right\}.$$ 
Based on Theorem \ref{th: rsc}, we have the following $\ell_1/\ell_2$ estimation error bounds.

\begin{theorem}[$\ell_1/\ell_2$ estimation error bounds of $\mathcal A$-UTrans]
	\label{th: error1}
	With the conditions on the regularizer $P_{\lambda}(\mb\beta)$ and conditions C1--C3, let $\lambda=c_1\sqrt{\frac{\log p}{n}}$ for a positive constant $c_1$. 
	Suppose $\mathcal A$ is known and $(s\log p/n)^{1/2}+h^{1/2}(\log p/n)^{1/4}=o(1)$,
	then there exists some positive constant $c$ such that 
	$$
	\|\wh{\mb\beta}_0-\mb\beta_0\|_{2} \lesssim \left(\frac{s\log p}{n}\right)^{1/2}+\left(\frac{\log p}{n}\right)^{1/4}h^{1/2}
	$$
	and
	$$
	\|\wh{\mb\beta}_0-\mb\beta_0\|_{1} \lesssim s\left(\frac{\log p}{n}\right)^{1/2}+\left(\frac{\log p}{n}\right)^{1/4}\left(sh\right)^{1/2}
	$$
	hold with probabilities at least $1-cp^{-1}$, where $h=\max_{k\in\mathcal A} \|\mb\beta_k-\mb\beta_0\|_1$.
\end{theorem}

Theorem \ref{th: error1} shows how the estimation errors of $\mb\beta_0$ are affected by $n_{\mathcal A}, n_0, s, p$, and $h$.
The $\ell_1$ error can be analyzed similarly to the $\ell_2$ error, so we only analyze the $\ell_2$ error here.
In transfer learning, we are more interested in the scenario of a small $n_0$ but diverging $n_{\mathcal A}$ since it is more realistic.
First, with a fixed $n_0$ and $n_{\mathcal A}\rightarrow\infty$, our result indicates that the estimation error goes to 0.
When the size of transferable source data is large enough, the effect of $h$ on estimation is dominated by an extremely large $n_{\mathcal A}$.
Indeed, as the simulation study shows (see Figure \ref{fig: detect} in Section \ref{sec: detect} and also Figure 3 in \cite{tian2022transfer}), the estimation error is dominated by a large $n_{\mathcal A}$ even with a relatively large $h$.
The scenarios of very large $h$ necessitate the source detection algorithm introduced in Section \ref{sec: detect}.
Besides, $h=0$ implies that the source data are completely transferable to the target data ($\mb\beta_k=\mb\beta_0$).
In this case, the $\ell_2$ error becomes $O_P(\sqrt{\frac{s\log p}{n}})$, the convergence rate of stacking all data vertically.
Second, without any available source data ($n_{\mathcal A}=0$ and $h=0$), the $\ell_2$ upper bound becomes $\sqrt{\frac{s\log p}{n_0}}$, the same rate as Lasso on target data only.
Third, Theorem \ref{th: error1} holds with the condition $s\log p=o(n_{\mathcal A})$ when $n_0\lesssim n_{\mathcal A}$, which is weaker than the condition $s\log p=o(n_0)$ for Lasso using the target data only.
Fourth, the $\ell_2$ error bound of $\mathcal A$-Trans-GLM (Theorem 1 of \cite{tian2022transfer}) is $(s\log p/n)^{1/2}+[(\log p/n_0)^{1/4}h^{1/2}]\wedge h$.
It is not hard to see that ours is the same as $\mathcal A$-Trans-GLM when $h\lesssim(\log p/n)^{1/2}$ and tighter than that when $h\gg(\log p/n)^{1/2}$ and $n_0 \ll n_{\mathcal A}$.

\begin{theorem}[Prediction error bound of $\mathcal A$-UTrans]
	\label{th: pred}
	Let $\mathcal E_{n_v}=1/n_v\|\mb X_v\left(\wh{\mb\beta}_0-\mb\beta_0\right)\|_{2}^2$ be the mean squared prediction error based on testing data $\mb X_v$.
	With the same conditions in Theorem \ref{th: error1} and some positive constant $c$, 
	\begin{equation*}
	 \mathcal E_{n_v}\lesssim \frac{s\log p}{n_v}+\left(\frac{\log p}{n_v}\right)^{3/4}\left(sh\right)^{1/2}+h\left(\frac{\log p}{n_v}\right)^{1/2}
	\end{equation*}
holds with probability at least $1-cp^{-1}$,
	where $\mb X_v$ is the testing data and $n_v$ is the corresponding testing data size.
\end{theorem}

\section{UTrans: Transfer Learning with Source Detection}
\label{sec: detect}

The $\mathcal A$-UTrans algorithm in Section \ref{sec: unified} assumes that the source data and the target data are similar to some extent, which might be unrealistic for an arbitrary dataset since $h$ can be small or large.
In fact, transferring nontransferable source data to the target data may bring adverse effects and lead to worse performance than the model with target data only \citep{pan2009survey, tian2022transfer}.
Therefore, a source detection algorithm is necessary in transfer learning.

Recall that our unified model, with $\mb X_k$ and $\mb X_0$, explicitly writes out the contrast $\mb\beta_k-\mb\beta_0$ with 
\begin{equation*}
\mb\mu=\begin{bmatrix}
		\mb X_k &\mb X_k\\
		\mb 0&\mb X_0\\
	\end{bmatrix} \begin{bmatrix}
	\mb\beta_k-\mb\beta_0 \\
	\mb\beta_0\\
\end{bmatrix}:=\mb W (\mb\beta_k-\mb\beta_0)+\mb Z \mb\beta_0
\end{equation*}
where $\mu=E(Y|\mb Z, \mb W)$, $\mb W=(\mb X_k^\top, \mb 0)^\top$, and $\mb Z=(\mb X_k^\top, \mb X_0^\top)^\top$.
Let $\mb\beta=[(\mb\beta_k-\mb\beta_0)^\top, \mb\beta_0^\top]^\top$ (note that $\mb\beta$ is defined differently from that in Section \ref{sec: unified}).
By testing $H_0: \mb\beta_k-\mb\beta_0=\mb 0$ vs $H_1: \mb\beta_k-\mb\beta_0\neq\mb 0$, we detect if the source data $\mb X_k$ are transferable to $\mb X_0$.

Both the parameter of interest $\mb\beta_k-\mb\beta_0$ and the nuisance parameter $\mb\beta_0$ are $p$-dimensional.
Methods on testing the high-dimensional vector with high-dimensional nuisance parameter is very limited in the literature. Recently, \citet{chen2022testing} proposes a U test statistic for the high-dimensional regression models, which extends the results of testing the low-dimensional parameter of interest in \citet{goeman2011testing} and \citet{guo2016tests}.
We propose an asymptotic $\alpha$-level test that rejects $H_0$ if 
$$|\hat{U}_{n_k}|/\sqrt{2\hat R_{n_k}}>z_{1-\alpha/2}$$
where $z_{1-\alpha/2}$ is the $(1-\alpha/2)$-th quantile of a standard normal distribution and
$$
	\hat{U}_{n_k}=\frac{1}{n_k} \sum_{i \neq i'}^{n_k}\left\{\left(y_i-\hat{\mu}_{\varnothing i}\right)\left(y_{i'}-\hat{\mu}_{\varnothing i'}\right) \mb x_{ki}^\top \mb x_{ki'}\right\} 
$$
$$\hat{R}_{n_k}=\frac{1}{n_k^2-n_k} \sum_{i \neq i'}^{n_k}\left\{\left(y_i-\hat{\mu}_{\varnothing i}\right)^2\left(y_{i'}-\hat{\mu}_{\varnothing i'}\right)^2\left(\mb x_{ki}^\top \mb x_{ki'}\right)^2\right\},$$
$\hat{\mu}_{\varnothing i}=\mb x_{ki}^\top \wh{\mb\beta}_0$ where $\wh{\mb\beta}_0$ is obtained by fitting $\mu=\mb z^\top\mb\beta_0$ under the null hypothesis.
Note that $\mb z$ is high-dimensional, so we obtain $\wh{\mb\beta}_0$ with the Lasso regression.
Denote $\Lambda_W^{\epsilon}=\text{tr}[E\{\text{var}(\epsilon)\mb x_k\mb x_k^\top\}^2]$ where $\epsilon=\mb y_k-\mb x_k^\top \mb\beta_k$.
Assume
\begin{itemize}
\item C4. Under $H_0$, there exist finite positive constants $c_1$ and $C_1$ such that 
$$c_1\leq \lambda_{\text{min}}\left\{E(\mb X_k\mb X_k^\top)\right\}\leq \lambda_{\text{max}}\left\{E(\mb X_k\mb X_k^\top)\right\}\leq C_1,$$
where $\lambda_{\text{min}}$ and $\lambda_{\text{max}}$ denote the smallest and largest eigenvalues of $E(\mb X_k\mb X_k^\top)$, respectively. 
\end{itemize}

\begin{theorem}
\label{th: test}
Assume the conditions C1--C2 and C4 and $s\log p/n=o(1)$.
Under $H_0$, if $n_k s\log p/(n \sqrt{2\Lambda_W^{\epsilon}})=o(1)$, then
$$\lim_{n_k\rightarrow\infty} \sup_{\|\mb\beta_0\|_2=O(1)}P\left(\frac{|\hat{U}_{n_k}|}{\sqrt{2\hat R_{n_k}}}>z_{1-\alpha/2}\right)=\alpha.$$
\end{theorem}

Theorem \ref{th: test} shows the probability of making the type I error (incorrectly excluding $\mb X_k$ when it is transferable).
Under some conditions, we find that the probability of making such error becomes small as $n_k\rightarrow\infty$.

\begin{algorithm}[hbt!]
	\caption{\textbf{UTrans}}
	\label{algorithm2}
	\SetAlgoLined
	\KwIn{$\{(\mb X_k, \mb y_k), 0\leq k\leq K'\}$.}
	
	\SetKwArray{fluxx}{$flux_x$}
	\SetKwArray{fluxy}{$flux_y$}
	\SetKwArray{q}{q}
	\For{$k \gets 1$  \KwTo $K'$} {%
    (1) write $\mb W=(\mb X_k^\top, \mb 0)^\top$ and $\mb Z=(\mb X_k^\top, \mb X_0^\top)^\top$.

    (2) estimate $\wh{\mb\beta}_0$ by fitting the model $\mu=\mb z^\top\mb\beta_0$.

    (3) compute $\hat{\mu}_{\varnothing i}=\mb x_{ki}^\top \wh{\mb\beta}_0$ and calculate the test statistic $t_k=|\hat{U}_{n_k}|/\sqrt{2\hat R_{n_k}}$.	
	}
 
$\wh{\mathcal A}=\left\{k: t_k\leq z_{1-\alpha/2/K'}\right\}$.
	
	\textbf{$\mathcal A$-UTrans:} obtain $\wh{\mb\beta}_0$ by minimizing (\ref{eq: loss}); obtain $\mathcal T_k=\{j: (\wh{\mb\beta_k-\mb\beta_0})_{j}=0\}.$
	
	\KwOut{$\wh{\mathcal A}$, $\wh{\mb\beta}_0$, and $\mathcal T_k$.}
\end{algorithm}

Algorithm UTrans
utilizes the tool of hypothesis testing to detect transferable source data.
To the best of our knowledge, this is the first work of using statistical inference for source detection in transfer learning.
We point out the benefit of our source detection algorithm. 
Compared to Trans-GLM \citep{tian2022transfer} which depends on the unknown constant $C_0$, our algorithm has no extra unknown parameters.
In fact, $C_0$ determines the threshold to select the transferable source data.
Without knowing the true value of $C_0$, a large value overestimates $\mathcal A$ and a small value underestimates $\mathcal A$.
Another round of cross validation can be run to find $C_0$, but increases computational cost.
Nevertheless, our algorithm estimates $\wh{\mathcal A}$ by directly testing $\mb\beta_k-\mb\beta_0=\mb 0$, which is more computationally efficient.

\section{Experiments}
\label{sec: simu}	
We illustrate the performances of our $\mathcal A$-UTrans and UTrans in various settings in terms of the averaged $\ell_2$-estimation error and the mean squared prediction error.
More specifically, we compare the following models: (1) $\mathcal A$-Trans-GLM and Trans-GLM: a two-step transferring model for linear regression without and with source detection, respectively, proposed by \citet{tian2022transfer}; 
(2) Trans-Lasso: a two-step transfer learning model for linear regression with source detection, proposed by \cite{Li2022};
(3) naive-Lasso: a model that fits the target data only using Lasso regression \citep{tibshirani1996regression}; (4) $\mathcal A$-UTrans-Lasso and $\mathcal A$-UTrans-SCAD: the proposed unified transfer learning models with Lasso and SCAD penalties.
R packages \textit{glmtrans} and \textit{glmnet} are used to implement Trans-GLM and Lasso on the target data only, respectively \citep{R2024}.
Our UTrans is implemented by the R package \textit{ncvreg}.

We consider 10 different settings for the number of source data, i.e., $K$ (subsection \ref{simu: normal}) and $K'$ (subsection \ref{simu: detect}) range from 1 to 10. 
With each $K$ and $K'$, experiments are replicated 200 times. 
Note that methods in \cite{Li2022} and \cite{tian2022transfer} mainly differ in the source detection algorithms, so we only compare Trans-Lasso in subsection  \ref{simu: detect}.

\subsection{Simulation with known $\mathcal A$}
\label{simu: normal}

This subsection is to show the theoretical properties in Theorem \ref{th: error1} and the advantages of our $\mathcal A$-UTrans algorithms in high-dimensional transfer learning with different dimensionalities $p$, target sizes $n_0$, and transferring levels $h$. 
We consider simulations with $n_0\in\{50, 75, 100\}$, $p\in\{300, 500, 600, 900\}$, and $h\in\{5, 10, 20, 40\}$.
We let the sample size of the source data $n_k=100$ for all $k=1,\cdots, K$ and fix the sparsity level $s=5$ in the target data.

For the target data, let $\mb\beta_0=(\mb{0.5}_s, \mb{0}_{p-s})$, where $\mb{0.5}_s$ means $s$ repetitions of 0.5 and $\mb{0}_{p-s}$ means $p-s$ repetitions of 0.
Each target sample $\mb x_{0i}\stackrel{iid}{\sim} \mathcal{N}(\mb 0_p, \mb\Sigma)$ with element $\Sigma_{jj'}=0.5^{|j-j'|}$ for $i=1,\cdots,n_0$ and $1\leq j, j'\leq p$.
For the $k$-th source data, we let $\mb\beta_k=(\mb{0.5}_s+(h/p)\mathcal{R}_s, \mb{0}_{p-s})$, where $\mathcal{R}_s$ is a $s$-dimensional independent Rademacher variable.
Each sample is generated from a $p$-dimensional $\mathcal{N}(\mb 0_p, \mb\Sigma+\mb\epsilon\mb\epsilon^\top)$ with $\mb\epsilon\sim \mathcal{N}(\mb 0_p, 0.3^2 \mb I_p)$.

\begin{figure}[hbt!]
	\centering
	\includegraphics[width=\linewidth,keepaspectratio]{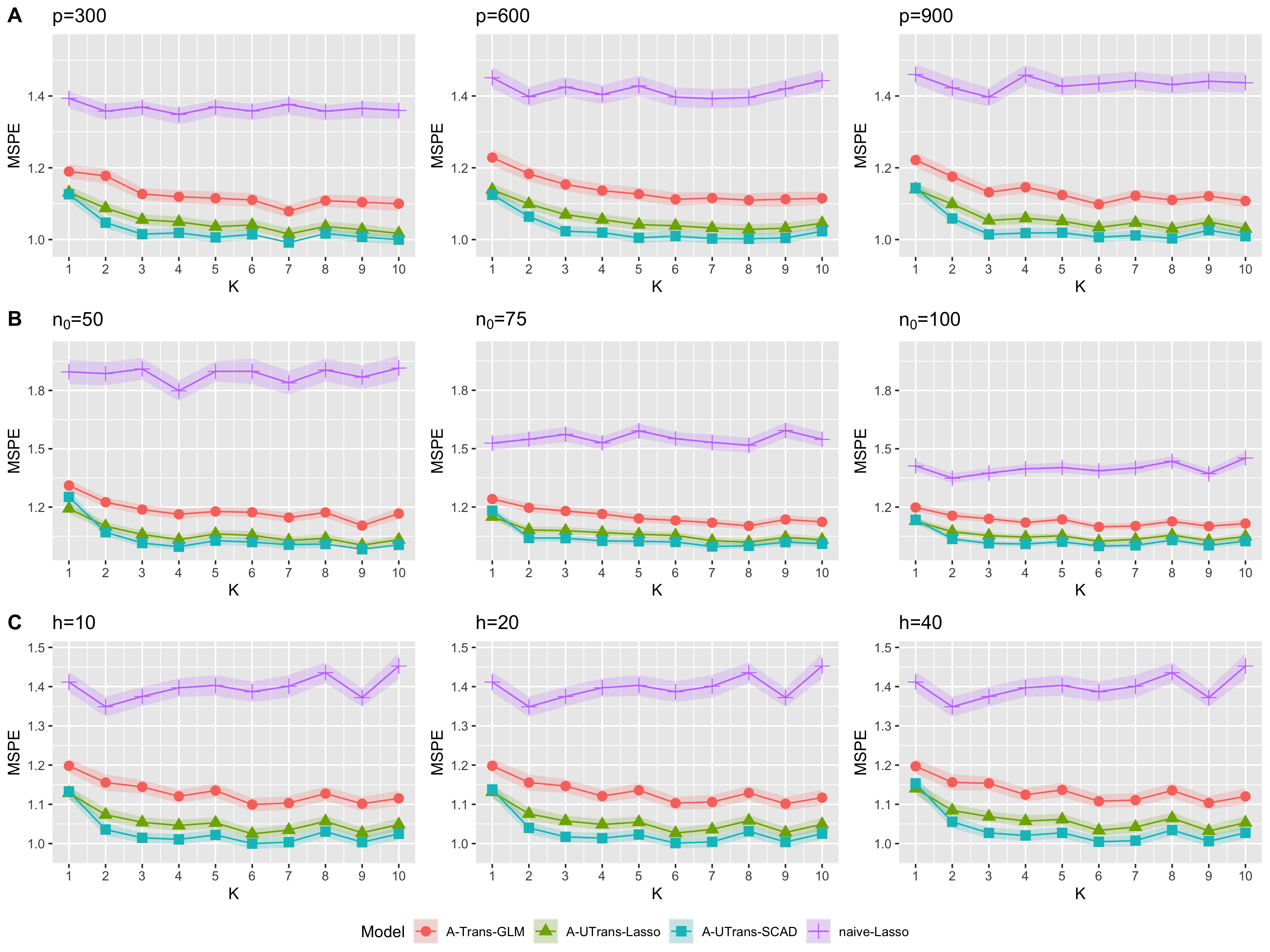}
	\caption{Mean squared prediction errors (MSPEs) of the proposed unified model and the existing transfer learning models with different settings of $p$ (row A), $n_0$ (row B), and $h$ (row C) for each $k=1,\cdots,K$. Shade areas are calculated by $\text{MSPE}\pm 0.1\times \text{standard deviation (SD)}$.}
	\label{fig: simu1}
\end{figure}

Figure \ref{fig: simu1} depicts the mean squared prediction errors of all models with different simulation settings.
More specifically, row A shows the results under different dimensionalities $p$.
We fix $n_0=100$ and $h=5$.
First, our proposed $\mathcal A$-UTrans-Lasso and $\mathcal A$-UTrans-SCAD outperform all the others.
Second, the naive-Lasso model fluctuates with the highest error no matter how $K$ increases, since $K$ controls the number of source data and naive-Lasso fits the target data only.
Row B shows the MSPEs of all models with different target sizes $n_0$.
We fix $p=500$ and $h=5$.
First, our proposed $\mathcal A$-UTrans algorithms have the best performances even with small sample sizes.
This evidence shows the benefit of transfer learning when the size of target data is small.
Second, the MSPEs of all models decrease as $n_0$ increases while $\mathcal A$-UTrans-SCAD attains the lowest error.
Row C illustrates the MSPEs of all models with various $h$.
We fix $n_0=100$ and $p=500$.
As the level $h$ increases, prediction errors of all transfer learning models with small $K$ increase but they fluctuate  as $K$ increases.

\begin{figure}[hbt!]
	\centering
	\includegraphics[width=\linewidth,keepaspectratio]{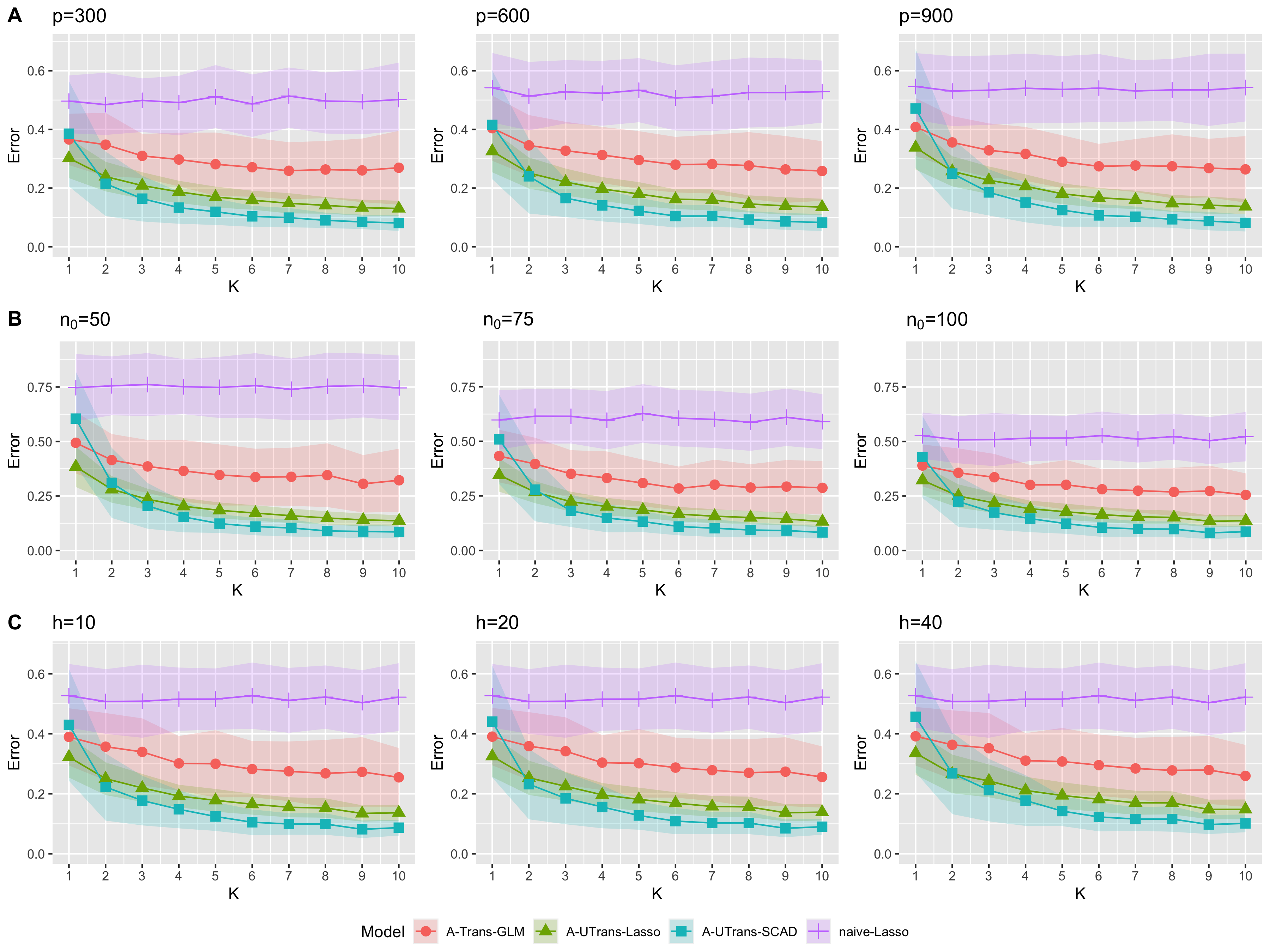}
	\caption{The averaged $\ell_2$ estimation errors of naive-Lasso, $\mathcal A$-Trans-GLM, $\mathcal A$-UTrans-Lasso, and $\mathcal A$-UTrans-SCAD with different settings. Shade areas are calculated by $\text{estimate}\pm \text{SD}$.}
	\label{fig: err1}
\end{figure}

Figure \ref{fig: err1} shows the averaged $\ell_2$ estimation errors of the four methods.
More specifically, our $\mathcal A$-UTrans algorithms obtain much lower errors than the others.
As $K$ increases, the errors of $\mathcal A$-UTrans-Lasso and $\mathcal A$-UTrans-SCAD drop dramatically.
This further shows that our algorithms have lower errors than the two-step $\mathcal A$-Trans-GLM.
The condition for improving the target model in \citet{Li2022} and \citet{tian2022transfer} allows $h$ as large as $\sqrt{n_0/\log p}$. 
In other words, they make a better upper bound better than the naive Lasso under this condition.
With these three simulation settings, this condition is satisfied in most settings and therefore improvement or theoretical property is granted.
While, our $\mathcal A$-UTrans still outperforms $\mathcal A$-Trans-GLM.
Overall, this simulation study presents that our proposed $\mathcal A$-UTrans maintains relatively low prediction errors in all settings. 
Particularly, $\mathcal A$-UTrans-SCAD outperforms all the others with relatively lower errors.

\subsection{Simulation with source detection}
\label{simu: detect}

In subsection \ref{simu: normal}, we consider the cases when the values of $h$ are relatively small.
Here, we consider the cases with relatively large $h$ and examine the effectiveness of the source detection algorithms.
We fix $p=500$, $K'=10$, and the source data sizes $n_k=200$ for $k\in\mathcal A$.
The target data are simulated in the same way as subsection \ref{simu: normal}.
For the $k$-th source data, each sample is generated from a $t$-distribution with degrees of freedom 4 and the covariance $\Sigma_{jj'}=0.5^{|j-j'|}$ for $i=1,\cdots,n_k$ and $1\leq j\neq j'\leq p$.
Note that we violate the assumptions C1 and C2 to show the robustness of UTrans with different data distributions.
We let $\mb\beta_0=(-\textbf{0.4}_3, -\textbf{0.5}_3, \textbf{0.6}_4, \textbf{0}_{490})$, $\mb\beta_k=\mb\beta_0$ if the $k$-th source data are transferable, and $\mb\beta_k=\mb\beta_0+h \mathcal{R}_p$ otherwise.

\begin{figure}[hbt!]
	\centering
	\includegraphics[width=\linewidth,keepaspectratio]{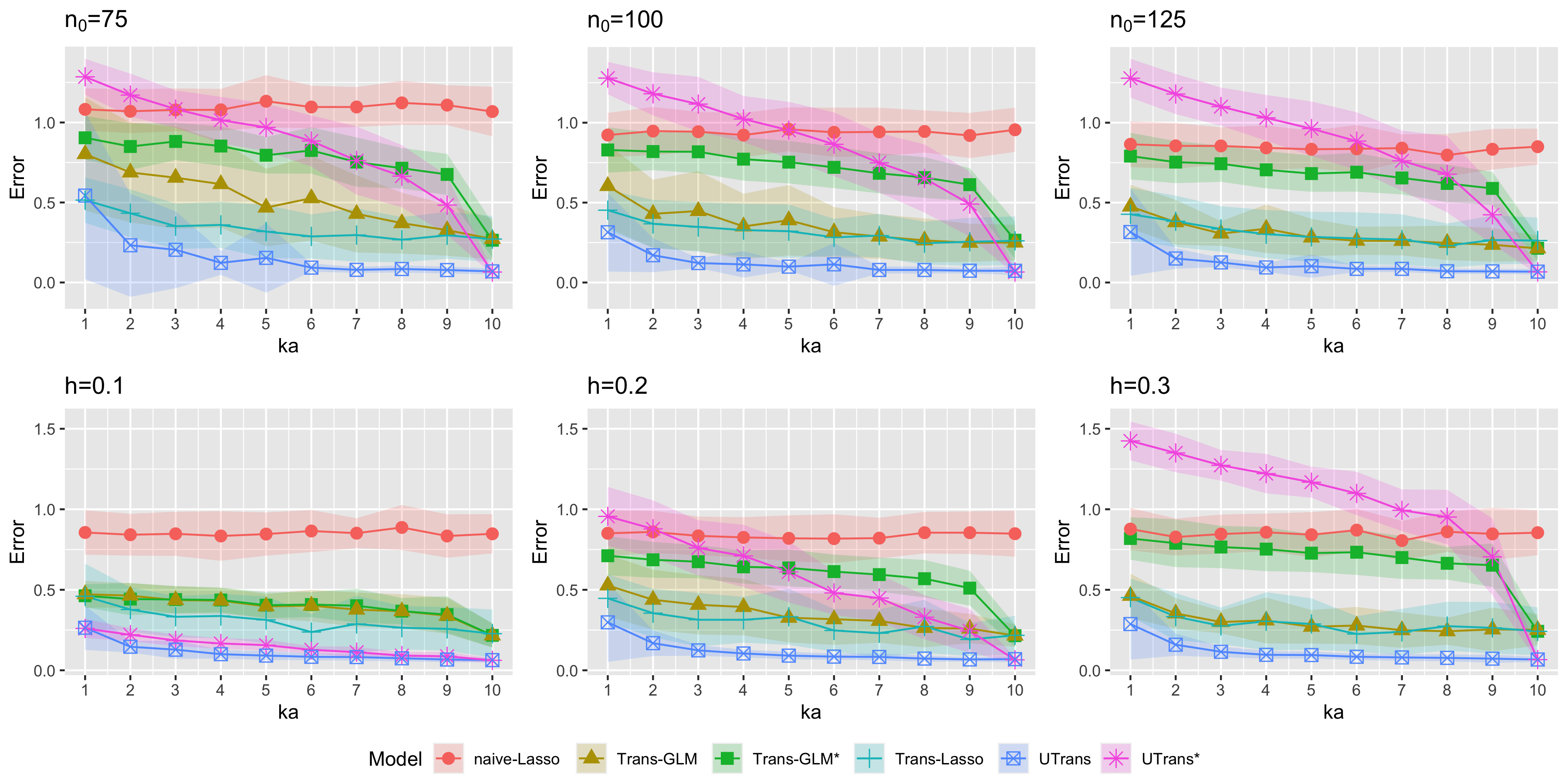}
	\caption{The averaged $\ell_2$ estimation errors of naive-Lasso, Trans-GLM, and UTrans with different settings.
		Shade areas are calculated by $\text{estimate}\pm \text{SD}$.}
	\label{fig: detect}
\end{figure}

\begin{figure}[hbt!]
	\centering
	\includegraphics[width=\linewidth,keepaspectratio]{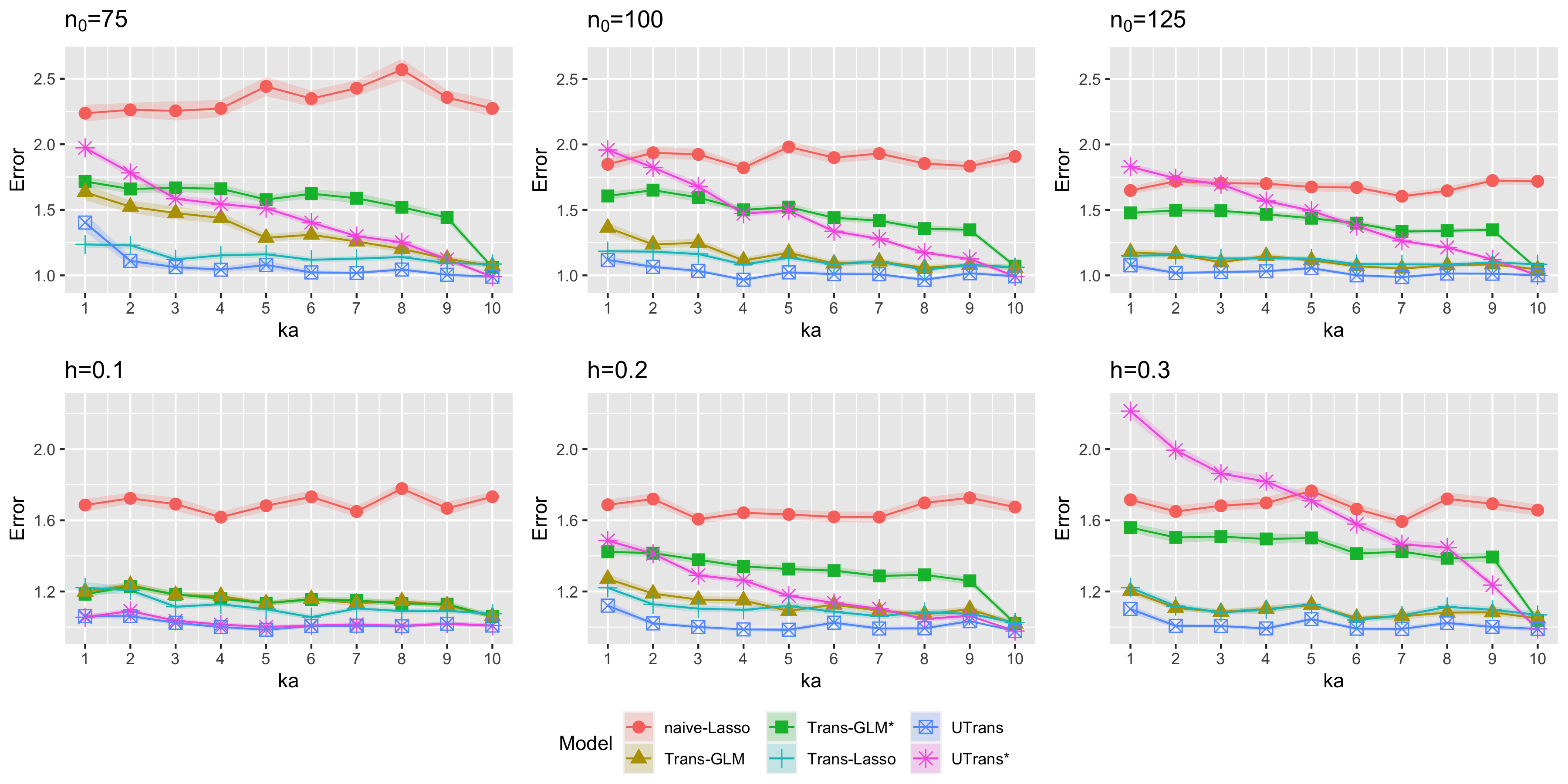}
	\caption{Mean squared prediction errors of the proposed unified model and the existing transfer learning models with different settings of $n_0$ and $h$. Shade areas are calculated by $\text{MSPE}\pm 0.1\times \text{SD}$.}
	\label{fig: d_pred}
\end{figure}

Figure \ref{fig: detect} and Figure \ref{fig: d_pred} show the estimation and prediction errors from naive-Lasso, Trans-GLM, Trans-GLM*, Trans-Lasso, UTrans, and UTrans*, respectively.
An algorithm with star denotes its pooled version, i.e., combining all the source data and target data. 
The x-axis ka represents the number of transferable source data.
The first row of Figure \ref{fig: detect} shows the estimation results with different target sizes and we fix $h=0.25$.
The second row demonstrates the results with different values of $h$ and we fix $n_0=100$.
When $n_0=75$, our algorithm UTrans obtains much lower estimation errors than Trans-GLM, which demonstrates the benefit of using transfer learning in the target data with relatively small size.
As $h$ increases, our UTrans keeps the lowest estimation errors among all algorithms in all settings, which shows the effectiveness of excluding nontransferable source data.
Overall, this study reveals that our proposed UTrans works better than existing algorithms in small target data and noisy source data.
Similar patterns for the prediction errors can be observed in Figure \ref{fig: d_pred}.

\section{Intergenerational Mobility Data}
\label{sec: app}

\begin{table*}[hbt!]
	\caption{The mean squared prediction errors for each target state. 
		Model names with stars are run on the pooled data.
		The bold numbers indicate the lowest prediction errors.}
	\centering
	\label{tab: state}
	\resizebox{\textwidth}{!}{\begin{tabular}{lrrrrrrrrrr}  \hline Model& AL & AR & CA&FL&LA&MN& NY&OK&PA&WI\\   \hline
RF&\textbf{4.6647} & 8.6112 & 0.4349 & \textbf{2.2448} & 1.1496 & 1.4110 & 0.4450 & 6.9399 & \textbf{0.5996} & 1.4255 \\
RF*&6.1231 & 7.7380 & 0.9830 & 2.3068 & 5.3322 & 1.7444 & 0.5937 & 7.2940 & 1.1103 & 1.7436 \\
XGBoost&6.2520 & 12.6309 & 0.6331 & 2.8952 & 1.5041 & 2.6424 & 0.6036 & 8.8445 & 0.8695 & 1.7809 \\
XGBoost*&5.9989 & 21.6605 & 0.6137 & 3.3179 & 4.7088 & 1.5407 & 0.7315 & 7.7890 & 27.9124 & 24.7909 \\
SVM&5.1180 & 7.5670 & 0.3473 & 2.4102 & 0.9202 & 1.1043 & 0.3923 & 6.3199 & 0.7585 & 1.2256 \\
SVM*&5.0375 & \textbf{7.5015} & 0.3339 & 2.3980 & \textbf{0.8894} & 1.2443 & \textbf{0.3791} & \textbf{6.2356} & 0.7465 & 1.2017 \\ 
Trans-Lasso&5.9565 & 9.6555 & 0.5729 & 2.4748 & 3.3234 & 1.4261 & 0.4930 & 7.0710 & 0.9106 & 1.4848 \\
Trans-GLM&5.4706 & 8.2521 & 0.4202 & 2.6550 & 0.9783 & 1.0620 & 0.3943 & 6.4283 & 0.9253 & 1.2637 \\
Trans-GLM*&5.5371 & 7.9981 & 0.4185 & 2.6622 & 0.9902 & 1.0621 & 1.0901 & 6.3438 & 0.8828 & 1.2772 \\
UTrans&5.0616 & 7.5426 & \textbf{0.3308} & 2.4154 & 0.8924 & \textbf{1.0566} & 0.3810 & 6.2586 & 0.7548 & \textbf{1.2011} \\
UTrans*&5.0572 & 7.5406 & 0.3328 & 2.4151 & 0.8924 & \textbf{1.0566} & 0.3811 & 6.2581 & 0.7549 & 1.2066 \\
\hline

\end{tabular}}
\end{table*}

\subsection{Data description}
We use the county-level data collected from the national census data, the Opportunity Atlas, and Data Commons to illustrate our UTrans.
Intergenerational mobility is measured as the change in income percentile for the children of all parents at the 75th national income percentile when they are aged 26.
Furthermore, we subset states with the numbers of counties larger than 50 for analysis. 
Of which, states with the numbers of counties between 50 and 75 are treated as the target states while others larger than 75 are treated as the source states.
Besides, we add two-way interactions of the county-level characteristics.
Overall, the processed data contain 1803 counties and 7875 predictors.
The states of interest (target states) include Alabama (AL-66), Arkansas (AR-64), California (CA-52), Florida (FL-65), Louisiana (LA-58), Minnesota (MN-69), New York (NY-61), Oklahoma (OK-60), Pennsylvania (PA-64), and Wisconsin (WI-68).
The source states include Georgia (GA-127), Illinois (IL-88), Indiana (IN-87), Iowa (IA-81), Kentucky (KY-98), Michigan (MI-76), Missouri (MO-87), North Carolina (NC-95), Ohio (OH-88), Tennessee (TN-86), Texas (TX-150), and Virginia (VA-113).
Number in the brackets denotes sample size, i.e., the number of counties.

\subsection{Predictive analysis}

We compare our UTrans to the following algorithms: Trans-GLM, Trans-GLM*, random forest (RF), RF*, XGBoost, XGBoost*, support vector machine (SVM), SVM*, UTrans, and UTrans*, where * denotes the pooled version, i.e., stacking both the source data and the target data.
We repeat our experiment 200 times and
evaluate these algorithms by the mean squared prediction error.
When applying these algorithms, we treat one state as the target data.
To make predictions, we randomly split 80\% of the target data as training data and the remaining 20\% as testing data.

Table \ref{tab: state} shows the mean squared prediction errors for each target state.
For each target state, the algorithm with the best performance is highlighted in bold.
Notably, UTrans performs the best in three states (CA, MN, and WI).
Compared to XGBoost and SVM and their pooled versions, UTrans still maintains relatively low prediction errors.
Compared to RF and RF*, which have the lowest errors in three states, our UTrans is also more interpretable in terms of variable importance.
The coefficients in linear regression represent the strength and direction of the relationship. RF is generally more difficult to interpret than linear regression since the individual trees can interact in complex ways and the importance of each feature may not be easily discernible from the output.
Compared to the transfer learning models Trans-Lasso and Trans-GLM, UTrans performs better than them in all the target states.

\section{Broader Impact}
\label{sec: dis}
In this paper, we propose a novel, unified, and interpretable transfer learning model with high-dimensional data.
To the best of our knowledge, this unified model is the first work on transfer learning that identifies both transferable variables and transferable source data;
It is also the first work that incorporates the statistical inference tool into transfer learning for source detection.
Multiple researches can be directed based on our framework.
First, our unified model may be extended to the nonlinear models with extra conditions, such as logistic regression, survival models, etc.
Second, our model will shed a light on the statistical learning community since it explicitly writes the contrasts.
Developing more powerful tools for source detection is very critical in transfer learning.

\newpage
\section*{Appendix}
The appendix contains technical proofs in Appendix A and additional simulation results in Appendix B.

\subsection*{Appendix A: Technical proof}

\begin{lemma}[Proposition 5.16 \citep{vershynin2010introduction}]
	\label{th: sub-exp}
	Let $x_1,\cdots,x_n$ be independent centered sub-exponential random variables, and let $M=\max_i\|x_i\|_{\psi_1}$. Then, for every $\mb a=(a_1,\cdots,a_n)^\top\in\mathbb{R}^n$ and every $t\geq 0$, we have 
	$$P\left(\left\Vert \sum_{i=1}^na_ix_i\right\Vert\geq t\right)\leq 2\exp\left[-c\min\left(\frac{t^2}{M^2\|\mb a\|_2^2},\frac{t}{M\|\mb a\|_{\infty}}\right)\right],$$
	where $c>0$ is an absolute constant.
\end{lemma}

\begin{lemma}[Lemmas 4(b) and 5 of \citet{Loh2015}]
	\label{th: loh}
	With the regularization function $P_{\lambda}$ satisfying the conditions (i)--(v),
	\begin{enumerate}
		\item For any $\mb w$, we have $\lambda L\|\mb w\|_1\leq P_{\lambda}(\mb w)+\tau/2 \|\mb w\|_2^2$
		\item Let $\mathcal{I}$ be the index set of the $s^*$ largest elements of $\mb v$ in magnitude.
		Suppose $\xi>0$ is such that $\xi P_{\lambda}(\mb v_{\mathcal I})- P_{\lambda}(\mb v_{\mathcal I^c})\geq0$, then
		$$\xi P_{\lambda}(\mb v_{\mathcal I})- P_{\lambda}(\mb v_{\mathcal I^c})\leq \lambda L\left(\xi\|\mb v_{\mathcal I}\|_1-\|\mb v_{\mathcal I^c}\|_1\right).$$
  Moreover, if $\mb\beta^*$ is $s^*$-sparse, then for an vector ${\mb\beta}$ such that $\xi P_{\lambda}(\mb\beta^*)-P_{\lambda}(\mb\beta)>0$ and $\xi\geq 1$, we have
  $$\xi P_{\lambda}(\mb\beta^*)-P_{\lambda}(\mb\beta)\leq \lambda L(\xi\|\mb v_{\mathcal I}\|_1-\|\mb v_{\mathcal I^c}\|_1)$$
  where $\mb v=\mb\beta-\mb\beta^*$.
	\end{enumerate}
\end{lemma}

\vskip4mm

\textbf{Proof of Theorem \ref{th: rsc}}
Denote $n=n_{\mathcal A}+n_0$.
First, it is not hard to derive
\begin{equation*}
	\begin{aligned}
		\wh{\mb\Sigma}=\frac1n \mb X^\top\mb X&=\frac1n\begin{bmatrix}
			\mb X^\top_1 & \mb 0 & \cdots&\cdots&\mb 0&\mb 0\\
			\mb 0 & \mb X^\top_2 & \mb 0&\cdots&\mb 0&\mb 0\\
			\vdots&\vdots&\vdots&\vdots&\vdots&\vdots\\
			\mb 0& \cdots&\cdots&\cdots&\mb X^\top_K&\mb 0\\
			\mb X^\top_1& \mb X^\top_2&\cdots&\cdots&\mb X^\top_K&\mb X^\top_0
		\end{bmatrix}     
		\begin{bmatrix}
			\mb X_1 & \mb 0 & \cdots&\cdots&\mb 0&\mb X_1\\
			\mb 0 & \mb X_2 & \mb 0&\cdots&\mb 0&\mb X_2\\
			\vdots&\vdots&\vdots&\vdots&\vdots&\vdots\\
			\mb 0& \cdots&\cdots&\cdots&\mb X_K&\mb X_K\\
			\mb 0& \cdots&\cdots&\cdots&\mb 0&\mb X_0
		\end{bmatrix}\\
		&=\frac1n\begin{bmatrix}
			n_1\wh{\mb\Sigma}_1 & \mb 0 & \cdots&\cdots&\mb 0&n_1\wh{\mb\Sigma}_1\\
			\mb 0 & n_2\wh{\mb\Sigma}_2 & \mb 0&\cdots&\mb 0&n_2\wh{\mb\Sigma}_2\\
			\vdots&\vdots&\vdots&\vdots&\vdots&\vdots\\
			\mb 0& \cdots&\cdots&\cdots&n_K\wh{\mb\Sigma}_K&n_K\wh{\mb\Sigma}_K\\
			n_1\wh{\mb\Sigma}_1& \cdots&\cdots&\cdots&n_K\wh{\mb\Sigma}_K&\sum_{k\in\mathcal{A}\cup\{0\}}n_k\wh{\mb\Sigma}_k\\
		\end{bmatrix}.
	\end{aligned}
\end{equation*}	

For any $\mb\Delta=(\mb\Delta_1^\top,\cdots,\mb\Delta_K^\top,\mb\Delta_0^\top)^\top$, we have
\begin{equation*}
	\begin{aligned}
		&\mb\Delta^\top\wh{\mb\Sigma}\mb\Delta=\begin{bmatrix}
			n_1/n\mb\Delta_1^\top\wh{\mb\Sigma}_1+n_1/n\mb\Delta_0^\top\wh{\mb\Sigma}_1\\
			\vdots\\
			n_K/n\mb\Delta_K^\top\wh{\mb\Sigma}_K+n_K/n\mb\Delta_0^\top\wh{\mb\Sigma}_K\\
			\sum_{k\in\mathcal{A}} n_k/n\mb\Delta_k^\top\wh{\mb\Sigma}_k+\sum_{k\in\mathcal{A}\cup\{0\}} n_k/n\mb\Delta_0^\top\wh{\mb\Sigma}_k
		\end{bmatrix}^\top
		\begin{bmatrix}
			\mb\Delta_1\\
			\vdots\\
			\mb\Delta_K\\
			\mb\Delta_0\\
		\end{bmatrix}\\
		&=\sum_{k\in\mathcal{A}}\frac{n_k}n\left\{  \mb\Delta_k^\top\wh{\mb\Sigma}_k\mb\Delta_k+2\mb\Delta_0^\top\wh{\mb\Sigma}_k\mb\Delta_k+\mb\Delta_0^\top\wh{\mb\Sigma}_k\mb\Delta_0 \right\}+\frac{n_0}n \mb\Delta_0^\top\wh{\mb\Sigma}_0\mb\Delta_0\\
		&=\sum_{k\in\mathcal{A}}\frac{1}n \left\Vert\mb X_k\mb\Delta_k+\mb X_k\mb\Delta_0\right\Vert_2^2+\frac{n_0}n \mb\Delta_0^\top\wh{\mb\Sigma}_0\mb\Delta_0\\
		&= \sum_{k\in\mathcal{A}}\frac{n_k}n \left(\mb\Delta_0+\mb\Delta_k\right)^\top\wh{\mb\Sigma}_k\left(\mb\Delta_0+\mb\Delta_k\right)+\frac{n_0}n \mb\Delta_0^\top\wh{\mb\Sigma}_0\mb\Delta_0\\
		&\geq \sum_{k\in\mathcal A} \left(v''\left\Vert\mb\Delta_k+\mb\Delta_0\right\Vert_2^2-\tau_k\sqrt{\frac{n_k\log p}{n^2}}\left\Vert\mb\Delta_k+\mb\Delta_0\right\Vert_1\right)+v'_0\|\mb\Delta_0\|_2^2-\tau_0 \sqrt{\frac{n_0\log p}{n^2}}\|\mb\Delta_0\|_1,
	\end{aligned}
\end{equation*}
where $v''=\min_k v_k n_k/n$, $v'=v''/2$, $v'_0=(2K+1)v'$, and the last inequality follows the RSC conditions on $\wh{\mb\Sigma}_k$.

In the context of our model, we replace $\mb\Delta$ by $\wh{\mb\Delta}=\wh{\mb\beta}-\mb\beta$. 
We observe
\begin{equation}
	\label{eq: l1p1}
	\begin{aligned}
		v'\|\wh{\mb\Delta}\|_2^2&=v'\sum_{k\in\mathcal A}\|\wh{\mb\Delta}_k\|^2+v'\|\wh{\mb\beta}_0-\mb\beta_0\|^2\\
		&= v'\sum_{k\in\mathcal A}\|\wh{\mb\Delta}_k+\wh{\mb\Delta}_0-\wh{\mb\Delta}_0\|^2+v'\|\wh{\mb\Delta}_0\|^2\\
&\leq 2v'\sum_{k\in\mathcal A}\left(\|\wh{\mb\Delta}_k+\wh{\mb\Delta}_0\|^2+\|\wh{\mb\Delta}_0\|^2\right)+v'\|\wh{\mb\Delta}_0\|^2\\
		&= v''\sum_{k\in\mathcal A} \|\wh{\mb\Delta}_k+\wh{\mb\Delta}_0\|_2^2+v'_0\|\wh{\mb\Delta}_0\|_2^2.
	\end{aligned}
\end{equation}

Let $\tau_k=\tau$ for $k\in\mathcal A$ and $\tau_0=\tau(K+1)$.
Then, we can also derive
\begin{equation}
	\label{eq: l1p2}
	\begin{aligned}
		&\tau\sum_{k\in\mathcal A}\sqrt{\frac{n_k\log p}{n^2}}\|\wh{\mb\Delta}_k+\wh{\mb\Delta}_0\|_1\leq \tau \sum_{k\in\mathcal A} \sqrt{\frac{n_m \log p}{n^2}}\left(\|\wh{\mb\Delta}_k\|_1+\|\wh{\mb\Delta}_0\|_1\right)\\
		=& \tau \sqrt{\frac{n_m\log p}{n^2}} \|\wh{\mb\Delta}\|_1+\tau \sqrt{\frac{n_m\log p}{n^2}}(K-1) \|\wh{\mb\Delta}_0\|_1\\
		\leq&\tau \sqrt{\frac{n_m\log p}{n^2}} \|\wh{\mb\Delta}\|_1+\tau K \sqrt{\frac{n_m\log p}{n^2}} \|\wh{\mb\Delta}\|_1=\tau_0 \sqrt{\frac{n_m\log p}{n^2}} \|\wh{\mb\Delta}\|_1
	\end{aligned}
\end{equation}

\begin{equation}
	\label{eq: l1p3}
	\tau\sqrt{\frac{n_0\log p}{n^2}}\|\wh{\mb\Delta}_0\|_1\leq \tau_0\sqrt{\frac{n_0\log p}{n^2}}\|\wh{\mb\Delta}\|_1.
\end{equation}

Finally, combining inequalities (\ref{eq: l1p1}), (\ref{eq: l1p2}), and (\ref{eq: l1p3}), we have 
$$
\wh{\mb\Delta}^\top\wh{\mb\Sigma}\wh{\mb\Delta}\geq v'\|\wh{\boldsymbol{\Delta}}\|_{2}^{2}-\tau_0 \left(\sqrt{\frac{n_m\log p}{n^2}}+\sqrt{\frac{n_0\log p}{n^2}}\right)\|\wh{\boldsymbol{\Delta}}\|_{1} \text { for } \wh{\mb\Delta} \in \mathbb{R}^{p^*} \text { and }\|\wh{\mb\Delta}\|_{1} \geq 1.
$$

According to Lemma 10 of \citet{liu2022multiple}, the aforementioned inequality with $\|\boldsymbol{\Delta}\|_{1} \geq 1$ actually implies 
$$
\wh{\mb\Delta}^\top\wh{\mb\Sigma}\wh{\mb\Delta}\geq v'\|\wh{\boldsymbol{\Delta}}\|_{2}^{2}-\tau_0 \left(\sqrt{\frac{n_m\log p}{n^2}}+\sqrt{\frac{n_0\log p}{n^2}}\right)\|\wh{\boldsymbol{\Delta}}\|_{1} \text { for } \wh{\mb\Delta} \in \mathbb{R}^{p^*}
$$
for a constant $\tau_0\geq 0$ and $v'>0$.

$\blacksquare$

\vskip5mm

\textbf{Proof of Theorem \ref{th: error1}}

First, the regularized loss function (\ref{eq: loss}) can be rewritten as 
$$\frac12\mb\beta^\top\widehat{\mb\Sigma}\mb\beta-\frac{1}{n}\mb y^\top \mb X\mb\beta+P_{\lambda}(\mb\beta).$$
Let $\wh{\mb\Delta}=\wh{\mb\beta}-\mb\beta$. The first-order condition implies that for any solution $\wh{\mb\beta}$ in the interior of the constraint set, $\wh{\mb\Sigma}\wh{\mb\beta}-\frac1n\mb X^\top\mb y +\nabla P_{\lambda}(\wh{\mb\beta})=\mb 0$ and therefore
\begin{equation}
	\label{eq: firstorder}
	\wh{\mb\Delta}^\top\wh{\mb\Sigma}\wh{\mb\beta}+\langle\nabla P_{\lambda}(\wh{\mb\beta})-\frac1n\mb X^\top\mb y, \wh{\mb\Delta}\rangle=0.
\end{equation}

For simplicity, we use $\tau$ for $\tau_0$.
The RSC condition on each $\wh{\mb\Sigma}_k$ from Theorem \ref{th: rsc} implies 
\begin{equation}
	\label{eq: rsc22}
	\wh{\mb\Delta}^\top\wh{\mb\Sigma}\wh{\mb\Delta}\geq v'\|\wh{\boldsymbol{\Delta}}\|_{2}^{2}-\tau \left(\sqrt{\frac{n_m\log p}{n^2}}+\sqrt{\frac{n_0\log p}{n^2}}\right)\|\wh{\boldsymbol{\Delta}}\|_{1}.
\end{equation}
Subtracting (\ref{eq: firstorder}) from (\ref{eq: rsc22}), we have
\begin{equation}
	\label{eq: first+rsc2}
	-\wh{\mb\Delta}^\top\wh{\mb\Sigma}{\mb\beta}-\langle\nabla P_{\lambda}(\wh{\mb\beta})-\frac1n\mb X^\top\mb y, \wh{\mb\Delta}\rangle\geq v'\|\wh{\boldsymbol{\Delta}}\|_{2}^{2}-\tau \left(\sqrt{\frac{n_m\log p}{n^2}}+\sqrt{\frac{n_0\log p}{n^2}}\right)\|\wh{\boldsymbol{\Delta}}\|_{1}.
\end{equation}
Since the function $P_{\tau, \lambda}(\mb w)=P_{\lambda}(\mb w)+\frac{\tau}2 \|\mb w\|_2^2$ is convex \citep{Loh2015, liu2022multiple},
\begin{equation}
	\label{eq: convex}
	-\langle\nabla P_{\lambda}(\wh{\mb\beta}), \wh{\mb\Delta}\rangle\leq P_{\lambda}(\mb\beta)-P_{\lambda}(\wh{\mb\beta})+\frac{\tau}2 \|\wh{\mb\Delta}\|_2^2.
\end{equation}
Combining (\ref{eq: first+rsc2}) and (\ref{eq: convex}), we have
\begin{equation*}
	\begin{aligned}
		&v'\|\wh{\boldsymbol{\Delta}}\|_{2}^{2}-\tau \left(\sqrt{\frac{n_m\log p}{n^2}}+\sqrt{\frac{n_0\log p}{n^2}}\right)\|\wh{\boldsymbol{\Delta}}\|_{1}\\
		\leq& -\wh{\mb\Delta}^\top \wh{\mb\Sigma}\mb\beta+\frac1n \mb X^\top\mb y \wh{\mb\Delta}+P_{\lambda}(\mb\beta)-P_{\lambda}(\wh{\mb\beta})+\tau/2\| \wh{\mb\Delta}\|_2^2
	\end{aligned}
\end{equation*}

\begin{equation*}
	\begin{aligned}
		v'\|\wh{\mb\Delta}\|_2^2-\tau/2\| \wh{\mb\Delta}\|_2^2&\leq P_{\lambda}(\mb\beta)-P_{\lambda}(\wh{\mb\beta})+\left(\left\Vert\wh{\mb\Sigma}\mb\beta-\frac1n \mb X^\top\mb y\right\Vert_{\infty}\right)\| \wh{\mb\Delta}\|_1\\
		&+\tau \left(\sqrt{\frac{n_m\log p}{n^2}}+\sqrt{\frac{n_0\log p}{n^2}}\right)\|\wh{\mb\Delta}\|_1
	\end{aligned}
\end{equation*}

\begin{equation*}
	\begin{aligned}
		&v'\|\wh{\mb\Delta}\|_2^2-\tau/2\| \wh{\mb\Delta}\|_2^2\\
		\leq& P_{\lambda}(\mb\beta)-P_{\lambda}(\wh{\mb\beta})+\left\{\left\Vert\wh{\mb\Sigma}\mb\beta-\frac1n \mb X^\top\mb y\right\Vert_{\infty}+\tau \left(\sqrt{\frac{n_m\log p}{n^2}}+\sqrt{\frac{n_0\log p}{n^2}}\right)\right\}\| \wh{\mb\Delta}\|_1
	\end{aligned}
\end{equation*}

Next, we only need to bound $\|\wh{\mb\Sigma}\mb\beta-\frac1n \mb X^\top\mb y\|_{\infty}$. Note that
\begin{equation*}
	\begin{aligned}
&\left\Vert\wh{\mb\Sigma}\mb\beta-\frac1n \mb X^\top\mb y\right\Vert_{\infty}= \left\Vert\frac1n \mb X^\top\mb \epsilon\right\Vert_{\infty}\\
		\leq& \left\Vert\frac2n \sum_{k\in \mathcal{A}}\mb X_k^\top \mb\epsilon_k\right\Vert_{\infty}+\left\Vert\frac1n \mb X_0^\top \mb\epsilon_0 \right\Vert_{\infty}\\
		\leq& c_1 \sqrt{\frac{n_{\mathcal A}\log p}{n^2}}+c_2 \sqrt{\frac{n_0\log p}{n^2}}
	\end{aligned}
\end{equation*}
for some constants $c_1$ and $c_2$ with probability at least $1-4p^{-1}$.
The last inequality follows the fact that the product of sub-Gaussian random variables is a sub-exponential random variable.
Therefore, $x_{ij}\epsilon_i$ is sub-exponential according to condition C2.
Using Lemma \ref{th: sub-exp} with $\mb a=[1,\cdots,1]^\top$, we have
\begin{equation*}
	\begin{aligned}
		P\left(\frac2{n}\left\Vert\sum_{k\in\mathcal{A}}\mb X_k^\top \mb\epsilon_k\right\Vert_{\infty}>t\right)\leq 2p \max_{j\leq p, k\in\mathcal{A}} \exp\left\{-c \min\left(\frac{n^2 t^2}{4M_k^2 n_{\mathcal A}},\frac{nt}{2M_k}\right)\right\},
	\end{aligned}
\end{equation*}
where $M_k=\max_{1\leq i\leq n_k} \|x_{ki}\|_{\psi_1}$.
With $\log p=o(n_{\mathcal A})$ and $t=c_1 \sqrt{n_{\mathcal A}\log p/n^2}$, we have 
$$P\left(\frac2{n}\left\Vert\sum_{k\in\mathcal{A}}\mb X_k^\top \mb\epsilon_k\right\Vert_{\infty}\leq c_1  \sqrt{\frac{n_{\mathcal A}\log p }{n^2}}\right)\geq 1-2p^{-1}$$
for some constant $c_1$.
Similarly, we have
$$P\left(\frac1{n}\left\Vert\mb X_0^\top \mb\epsilon_0\right\Vert_{\infty}\leq c_2 \sqrt{\frac{n_0\log p}{n^2}}\right)\geq 1-2p^{-1}$$
for some constant $c_2$.
The last inequality follows by combining the aforementioned two inequalities such that
$$\left\Vert\wh{\mb\Sigma}\mb\beta-\frac1n \mb X^\top\mb y\right\Vert_{\infty}\leq c_1 \sqrt{\frac{n_{\mathcal A}\log p}{n^2}}+c_2 \sqrt{\frac{n_0\log p}{n^2}}\asymp \sqrt{\frac{\log p}{n}}$$
with probability at least $1-4p^{-1}$.
Then 
$$\left\Vert\wh{\mb\Sigma}\mb\beta-\frac1n \mb X^\top\mb y\right\Vert_{\infty}+\tau\left(\sqrt{\frac{n_m\log p}{n^2}}+\sqrt{\frac{n_0\log p}{n^2}}\right) \leq c_1 \sqrt{\frac{\log p}{n}},$$
for large enough $c_1$.

Let $\lambda= 2c_1 \sqrt{\frac{\log p}{n}}$, we have
\begin{equation*}
	\begin{aligned}
		v'\|\wh{\mb\Delta}\|_2^2-\tau/2\| \wh{\mb\Delta}\|_2^2&\leq P_{\lambda}(\mb\beta)-P_{\lambda}(\wh{\mb\beta})+\lambda/2\| \wh{\mb\Delta}\|_1\\
		&\leq P_{\lambda}(\mb\beta)-P_{\lambda}(\wh{\mb\beta})+1/2P_{\lambda}(\wh{\mb\Delta})+\tau/4 \|\wh{\mb\Delta}\|_2^2\\
  &\leq P_{\lambda}(\mb\beta)-P_{\lambda}(\wh{\mb\beta})+1/2P_{\lambda}(\mb\beta)+1/2 P_{\lambda}(\wh{\mb\beta})+\tau/4 \|\wh{\mb\Delta}\|_2^2,
	\end{aligned}
\end{equation*}
where the second inequality follows Lemma \ref{th: loh}.
With the second inequality in Lemma \ref{th: loh}, we finally have
$$2v'\|\wh{\mb\Delta}\|_2^2-3\tau/2\| \wh{\mb\Delta}\|_2^2\leq 3 \lambda L\|\wh{\mb\Delta}_{\mathcal I}\|_1-\lambda L \|\wh{\mb\Delta}_{\mathcal I^c}\|_1.$$

Besides,
\begin{equation}
	\label{eq: delta}
	\begin{aligned}
		&\|\wh{\mb\Delta}_{\mathcal I^c}\|_1=\sum_{k\in\mathcal A}\left\|\left[\wh{\mb\beta_k-\mb\beta_0}-(\mb\beta_k-\mb\beta_0)\right]_{\mathcal I^c}\right\|_1+\left\|(\wh{\mb\beta}_0-{\mb\beta}_0)_{\mathcal I^c}\right\|_1\\
		\geq & \sum_{k\in\mathcal A}\left\|(\wh{\mb\beta_k-\mb\beta_0)}_{\mathcal I^c}\right\|_1-\sum_{k\in\mathcal A}\left\|(\mb\beta_k-\mb\beta_0)_{\mathcal I^c}\right\|_1+\left\|(\wh{\mb\beta}_0-{\mb\beta}_0)_{\mathcal I^c}\right\|_1\\
		\geq & \sum_{k\in\mathcal A}\left\|(\wh{\mb\beta_k-\mb\beta_0)}_{\mathcal I^c}\right\|_1-Kh+\left\|(\wh{\mb\beta}_0-{\mb\beta}_0)_{\mathcal I^c}\right\|_1,
	\end{aligned}
\end{equation}
which implies
\begin{equation}
	\label{eq: deltaI}
	-\lambda L\|\wh{\mb\Delta}_{\mathcal I^c}\|_1\leq -\lambda L\sum_{k\in\mathcal A}\left\|(\wh{\mb\beta_k-\mb\beta_0)}_{\mathcal I^c}\right\|_1+\lambda LKh-\lambda L\left\|(\wh{\mb\beta}_0-{\mb\beta}_0)_{\mathcal I^c}\right\|_1.
\end{equation}

With Theorem \ref{th: rsc}, Eq. (\ref{eq: delta}), and Eq. (\ref{eq: deltaI}), we obtain
\begin{equation*}
	\begin{aligned}
		&2v'\|\wh{\mb\Delta}\|_2^2-3\tau/2\| \wh{\mb\Delta}\|_2^2\leq 3 \lambda L\|\wh{\mb\Delta}_{\mathcal I}\|_1-\lambda L \|\wh{\mb\Delta}_{\mathcal I^c}\|_1\\
		\leq & 3 \lambda L\|\wh{\mb\Delta}_{\mathcal I}\|_1+\lambda LKh\\
		\lesssim & 3\lambda \sqrt{s}\|\wh{\mb\Delta}\|_2+\lambda h.
	\end{aligned}
\end{equation*}

Let $a=2v'-3\tau/2$ for simplicity.
We have
\begin{equation*}
a\|\wh{\mb\Delta}\|_2^2\lesssim 3\lambda \sqrt{s}\|\wh{\mb\Delta}\|_2+ \lambda h.
\end{equation*}
Let $x=\|\wh{\mb\Delta}\|_2$, then we solve the quadratic inequality $ax^2-3\lambda \sqrt{s} x-\lambda h\lesssim 0$ and we have
$$\|\wh{\mb\Delta}\|_2\lesssim \lambda\sqrt{s}+\sqrt{\lambda h}.$$
Plugging in the choice of $\lambda$, we have 
\begin{equation*}
	\begin{aligned}
		\|\wh{\mb\Delta}\|_2
		&\lesssim \sqrt{\frac{s\log p}{n}}+\left(\frac{\log p}{n}\right)^{1/4}\sqrt{h}.
	\end{aligned}
\end{equation*}
Since $\wh{\mb\beta}_0-\mb\beta_0$ is a subset of $\wh{\mb\Delta}$, this result also holds for $\|\wh{\mb\beta}_0-\mb\beta_0\|_2$, i.e.,
\begin{equation*}
	\begin{aligned}
		\|\wh{\mb\beta}_0-\mb\beta_0\|_2\
		&\lesssim \sqrt{\frac{s\log p}{n}}+\left(\frac{\log p}{n}\right)^{1/4}\sqrt{h}.
	\end{aligned}
\end{equation*}

Immediately from the $\ell_2$ error of $\|\wh{\mb\beta}_0-\mb\beta_0\|_2$, we have
$$\|\wh{\mb\beta}_0-\mb\beta_0\|_1\lesssim s\sqrt{\frac{\log p}{n}}+\left(\frac{\log p}{n}\right)^{1/4}\sqrt{sh}.$$

$\blacksquare$
\vskip5mm

\textbf{Proof of Theorem \ref{th: pred}}

For simplicity, we drop the subscript $v$ in the testing data $(\mb X_v, \mb y_v)$.
Let $\mathcal L_n(\mb\beta)=\frac1{2n}\|\mb y-\mb X\mb\beta\|_2^2$ and $\wh{\mb\Delta}_0=\wh{\mb\beta}_0-\mb\beta_0$, then the prediction error is $$\langle\nabla\mathcal L_n(\wh{\mb\beta}_0)-\nabla\mathcal L_n(\mb\beta_0), \wh{\mb\Delta}_0\rangle=\frac1n\left\|\mb X(\wh{\mb\beta}_0-\mb\beta_0)\right\|_2^2=(\wh{\mb\beta}_0-\mb\beta_0)^\top \wh{\mb\Sigma}(\wh{\mb\beta}_0-\mb\beta_0)=\wh{\mb\Delta}_0^\top\wh{\mb\Sigma}\wh{\mb\Delta}_0.$$
Assume the RSC condition on the test data such that
$\mb\Delta^\top\wh{\mb\Sigma}\mb\Delta\geq v\|\mb\Delta\|_2^2-\tau\sqrt{\log p/n}\|\mb\Delta\|_1$ for any $\mb\Delta\in\mathbb{R}^p$.
Similar to the proof of Theorem \ref{th: error1}, we have
$$-\langle\nabla P_{\lambda}(\wh{\mb\beta}_0),\wh{\mb\Delta}_0\rangle\leq P_{\lambda}(\mb\beta_0)-P_{\lambda}(\wh{\mb\beta}_0)+\tau/2\|\wh{\mb\Delta}_0\|_2^2.$$
The first-order condition implies
$$\langle\nabla \mathcal L_n(\wh{\mb\beta}_0)+\nabla P_{\lambda}(\wh{\mb\beta}_0),-\wh{\mb\Delta}_0\rangle\geq 0.$$
Therefore, the prediction error
\begin{equation*}
	\begin{aligned}
		&\langle\nabla\mathcal L_n(\wh{\mb\beta}_0)-\nabla\mathcal L_n(\mb\beta_0), \wh{\mb\Delta}_0\rangle\leq \langle-\nabla \mathcal L_n(\mb\beta_0)-\nabla P_{\lambda}(\wh{\mb\beta}_0),\wh{\mb\Delta}_0\rangle\\
		\leq & P_{\lambda}(\mb\beta_0)-P_{\lambda}(\wh{\mb\beta}_0)+\tau/2\|\wh{\mb\Delta}_0\|_2^2+\|\nabla\mathcal L_n(\mb\beta_0)\|_{\infty}\|\wh{\mb\Delta}_0\|_1.
	\end{aligned}
\end{equation*}

Let $\mathcal M$ be the support set of $\mb\beta$, i.e., $\mathcal M=\{j: \beta_j\neq 0\}$.
Next, we bound $P_{\lambda}(\mb\beta_0)-P_{\lambda}(\wh{\mb\beta}_0)$ by
\begin{equation*}
	\begin{aligned}
		&P_{\lambda}(\mb\beta_0)-P_{\lambda}(\wh{\mb\beta}_0)=P_{\lambda}(\mb\beta_0)-P_{\lambda}(\wh{\mb\beta}_{0\mathcal M})-P_{\lambda}(\wh{\mb\beta}_{0\mathcal M^c})\\
		\leq &P_{\lambda}(\wh{\mb\Delta}_{0\mathcal M})-P_{\lambda}(\wh{\mb\beta}_{0\mathcal M^c})\\
		=&P_{\lambda}(\wh{\mb\Delta}_{0\mathcal M})-P_{\lambda}(\wh{\mb\Delta}_{0\mathcal M^c})\\
		\leq & \lambda L(\|\wh{\mb\Delta}_{0\mathcal M}\|_1-\|\wh{\mb\Delta}_{0\mathcal M^c}\|_1)\\
		\leq & \lambda L\|\wh{\mb\Delta}_{0}\|_1.
	\end{aligned}
\end{equation*}
Together with the result $\|\nabla\mathcal L_n(\mb\beta_0)\|_{\infty}\lesssim\lambda$ (from the proof of Theorem \ref{th: error1} or \citet{Loh2015}), we have
\begin{equation*}
	\begin{aligned}
		&\langle\nabla\mathcal L_n(\wh{\mb\beta}_0)-\nabla\mathcal L_n(\mb\beta_0), \wh{\mb\Delta}_0\rangle\\
		\lesssim& \lambda L \|\wh{\mb\Delta}_{0}\|_1+\frac{\tau}{2}\|\wh{\mb\Delta}_{0}\|_2^2+\lambda \|\wh{\mb\Delta}_{0}\|_1\\
		\lesssim & \lambda\sqrt{s}\|\wh{\mb\Delta}_{0}\|_2+\|\wh{\mb\Delta}_{0}\|_2^2.
	\end{aligned}
\end{equation*}

The result follows by plugging in the $\ell_2$ error bound in Theorem \ref{th: error1} such that
$$\frac1n\left\|\mb X(\wh{\mb\beta}_0-\mb\beta_0)\right\|_2^2\lesssim \frac{s\log p}{n}+\left(\frac{\log p}{n}\right)^{3/4}\sqrt{sh}+h\sqrt{\frac{\log p}{n}}.$$

$\blacksquare$

\textbf{Proof of Theorem \ref{th: test}}

We decompose $\hat{U}_{n_k}$ by
\begin{equation*}
\begin{aligned}
\hat{U}_{n_k}=&\underbrace{\frac{1}{n_k} \sum_{i \neq i'}^{n_k}\left\{\left(y_i-{\mu}_i\right)\left(y_{i'}-{\mu}_{i'}\right) \mb x_{ki}^\top \mb x_{ki'}\right\}}_\text{$I_{\hat U_{n_k}}$}+\underbrace{\frac{1}{n_k} \sum_{i \neq i'}^{n_k}\left\{\left(\mu_i-\hat{\mu}_{\varnothing i}\right)\left(\mu_{i'}-\hat{\mu}_{\varnothing i'}\right) \mb x_{ki}^\top \mb x_{ki'}\right\}}_\text{$II_{\hat U_{n_k}}$}\\
+&\underbrace{\frac{2}{n_k} \sum_{i \neq i'}^{n_k}\left\{\left(y_i-{\mu}_i\right)\left(y_{i'}-\hat{\mu}_{\varnothing i'}\right) \mb x_{ki}^\top \mb x_{ki'}\right\}}_\text{$III_{\hat U_{n_k}}$}.
\end{aligned}
\end{equation*}

Note that the size is proved under $H_0$, 
We first exam $II_{\hat U_{n_k}}$: note that 
$$\frac{II_{\hat U_{n_k}}}{n_k}=\underbrace{(\wh{\mb\beta}_0-\mb\beta_0)^\top\left[\frac1{n_k}\sum_{i=1}^{n_k}\mb z_i\mb w_i^\top\right]\left[\frac1{n_k}\sum_{i=1}^{n_k}\mb z_i\mb w_i^\top\right](\wh{\mb\beta}_0-\mb\beta_0)}_\text{II1}-\underbrace{\frac1{n_k^2}\sum_{i=1}^{n_k} (\mu_i-\hat{\mu}_{\varnothing i})^2 \mb w_i^\top\mb w_i}_\text{II2}.$$
For II1, let $\wh{\mb\Sigma}=\frac1{n_k}\sum_{i=1}^{n_k}\mb z_i\mb w_i^\top=\frac1{n_k}\sum_{i=1}^{n_k}\mb x_{ki}\mb x_{ki}^\top$ and $\mb\Sigma=E(\mb x_{ki}\mb x_{ki}^\top)$. 
Then, it can be shown that 
$$\|\wh{\mb\Sigma}-\mb\Sigma\|_{\infty}=\tau=O_p\left(\sqrt{\frac{\log p}{n_k}}\right).$$
Similar to A2 in \citet{chen2022testing}, we see that $|II1|=O_p(\|\wh{\mb\beta}_0-\mb\beta_0\|_2^2)=O_p\left(\frac{s\log p}{n}\right)$, where $n=n_0+n_k$.

For II2, $n_k II2\leq \|\mu-\hat{\mu}_{\varnothing}\|_{\infty}^2\frac1{n_k}\sum_{i=1}^{n_k} \mb x_{ki}^\top \mb x_{ki}=o_p(\sqrt{2\Lambda_W^{\epsilon}})$ .

Finally, $$II_{\hat U_{n_k}}=n_k II1+n_k II2=O_p(n_k s\log p/n)+o_p(\sqrt{2\Lambda_W^{\epsilon}})=o_p(\sqrt{2\Lambda_W^{\epsilon}})$$ 
when $n_k s\log p/n/\sqrt{2\Lambda_W^{\epsilon}}=o(1)$.

We next exam $III_{\hat U_{n_k}}$: Similar to \citet{chen2022testing}, we obtain $|III1|=O_p[\frac1{\sqrt{n n_k}}\sqrt{s\log p}(2\Lambda_W^{\epsilon})^{1/4}]$ and $n_k III2=o_p(\sqrt{2\Lambda_W^{\epsilon}})$.
Finally, 
$$III_{\hat U_{n_k}}=n_k III1+n_k III2=O_p\left[\sqrt{\frac{n_k s\log p}{n}}(2\Lambda_W^{\epsilon})^{1/4}\right]+o_p\left(\sqrt{2\Lambda_W^{\epsilon}}\right)=o_p\left(\sqrt{2\Lambda_W^{\epsilon}}\right)$$
when $n_k s\log p/n/\sqrt{2\Lambda_W^{\epsilon}}=o(1)$.
Remaining steps are the same as \citet{chen2022testing}.

$\blacksquare$

\subsection*{Appendix B: Additional simulation results}
\label{simu: other}

\begin{figure}[hbt!]
	\centering
	\includegraphics[width=\textwidth,keepaspectratio]{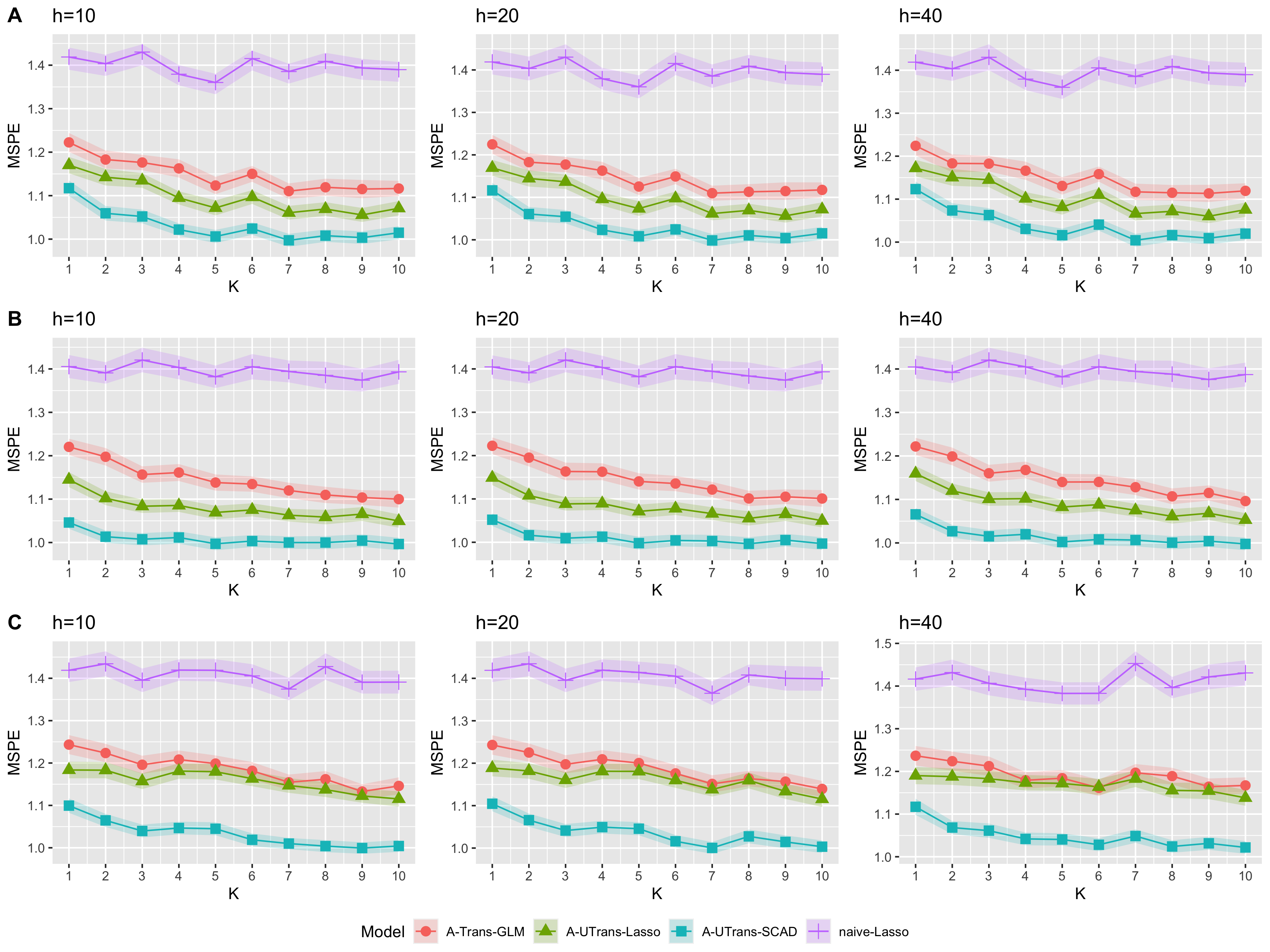}
	\caption{Mean squared prediction errors (MSPEs) of the proposed unified model and the existing transfer learning models with the compound symmetry (row A), $t$-distribution (row B), and Gaussian mixture model (row C) for each $k=1,\cdots,K$. Shade areas are calculated by $\text{MSPE}\pm 0.1\times \text{SD}$.}
	\label{fig: other}
\end{figure}

Setting 2: we consider design matrices of the source data from other distributions.
We fix the number of source data to $K=10$.
The sample sizes of the target data and all source data are 100.
The dimensionality $p=500$.
For the target data, $\mb x_0$ and $\mb\beta_0$ are simulated the same as those in subsection \ref{simu: normal}.
For the source data, $\mb\beta_k$ is simulated the same as the one in subsection \ref{simu: normal}, but the $i$-th data point in the $k$-th source data $\mb x_{ki}$ is simulated differently.
Specifically, we simulate the data points in the source data with three different distributions:
\begin{enumerate}
	\item $\mb x_{ki}\sim \mathcal{N}(\mb 0, \mb\Sigma)$ with  $\Sigma_{jj}=1$ and $\Sigma_{jj'}=0.5$. 
	\item $\mb x_{ki}$ follows a $t$-distribution with degrees of freedom 4.
	\item Data points in each source data are simulated from $0.5 \mathcal{N}(\frac{2\sqrt{5}}{5}, \frac15)+0.5 \mathcal{N}(-\frac{2\sqrt{5}}{5}, \frac15)$, a bimodal Gaussian mixture model.
	Note that the mixture model has mean 0 and variance 1.
\end{enumerate}
Note that the normality assumptions for the source data do not hold in the second and third cases.
Thus, these two cases are to exam the performance of UTrans under the nonnormal designs.

\begin{figure}[h]
	\centering
	\includegraphics[width=\linewidth,keepaspectratio]{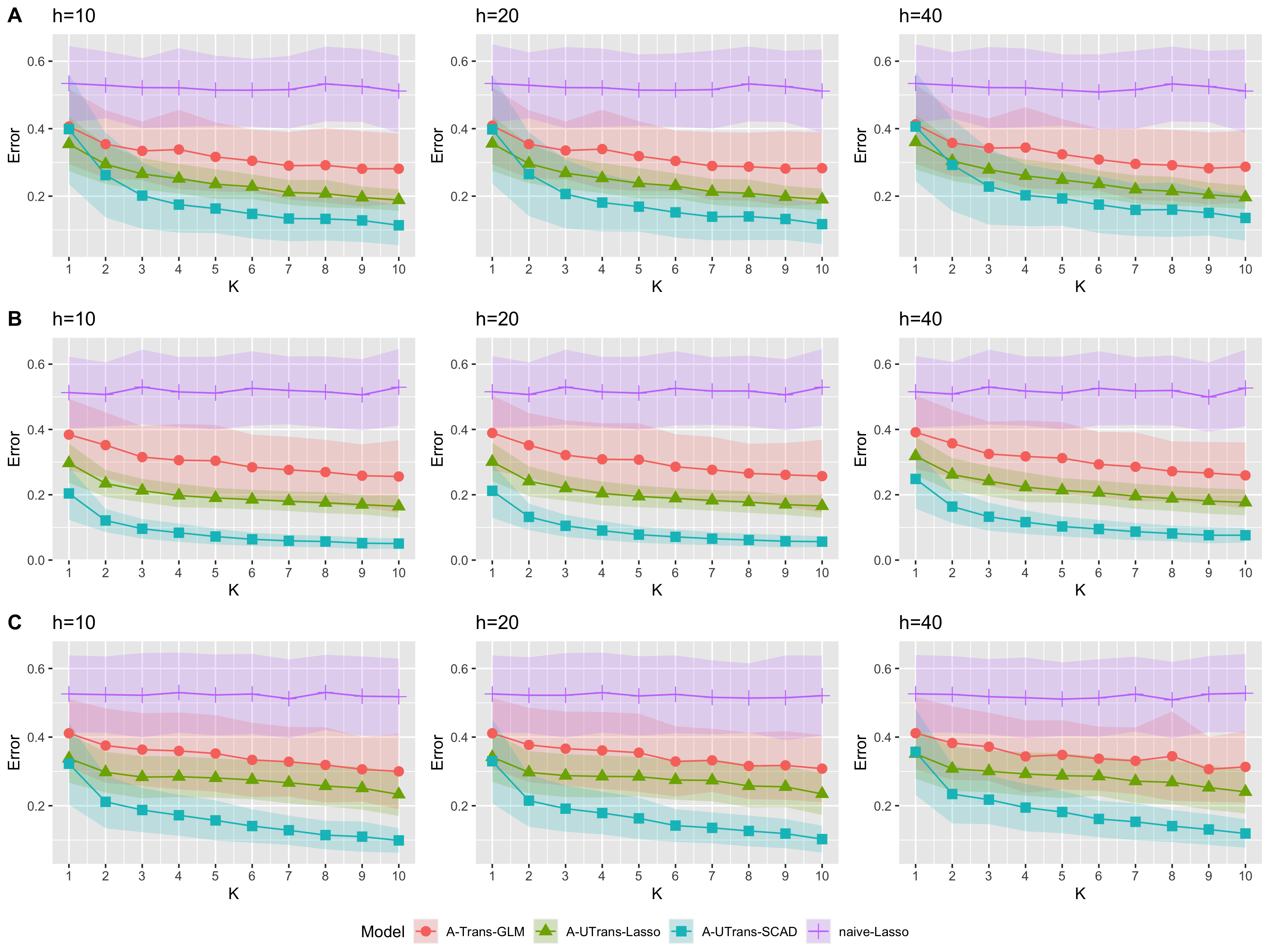}
	\caption{The averaged $\ell_2$ estimation errors of naive-Lasso, $\mathcal A$-Trans-GLM, $\mathcal A$-UTrans-Lasso, and $\mathcal A$-UTrans-SCAD in setting 2. Shade areas are calculated by $\text{estimate}\pm \text{SD}$.}
	\label{fig: err2}
\end{figure}

Row A shows the results of the simulated data under a normal distribution with the covariance structure of compound symmetry.
Compared to the existing $\mathcal A$-Trans-GLM, our proposed $\mathcal A$-UTrans algorithms attain the lowest prediction errors with various $h$.
Particularly, $\mathcal A$-UTrans-SCAD keeps the lowest errors all the time.
Row B presents the results when the source data are from the $t$-distribution and Row C illustrates the results when the source data are from a Gaussian mixture model.
$\mathcal A$-UTrans-Lasso performs similarly to $\mathcal A$-Trans-GLM while $\mathcal A$-UTrans-SCAD outperforms the others.
We observe that the prediction errors of $\mathcal A$-UTrans-SCAD are always the lowest among the others. 
In summary, Figure \ref{fig: other} demonstrates that our proposed method, particularly $\mathcal A$-UTrans-SCAD, outperforms the others when the data are simulated with more complicated structures.
Therefore, our method is more robust to the distributions of source data.
The $\ell_2$ estimation errors in these setting show similar patterns as Figure \ref{fig: err2}.

\newpage
\bibliography{ref}{}

\end{document}